\documentclass[twocolumn]{autart}

\usepackage{amsmath}
\usepackage{graphicx}
\usepackage[dvipsnames]{xcolor}
\usepackage{subfigure}
\usepackage{booktabs} %
\usepackage{nicefrac}       %

\usepackage{dsfont}
\usepackage{multirow} %

\usepackage{soul}
\usepackage{cancel}
\usepackage{graphicx}
\usepackage{algorithm}
\usepackage[]{algcompatible} %
\algnewcommand{\IIf}[1]{\State\algorithmicif\ #1\ \algorithmicthen}
\algnewcommand{\IElse}[1]{\State\algorithmicelse\ #1\ }
\algnewcommand{\ENDIIf}{\unskip\ \algorithmicend\ \algorithmicif}

\usepackage{xcolor}
\usepackage{amsmath,amssymb,amsfonts,bm}
\usepackage[shortlabels]{enumitem}
\setitemize{nosep}

\usepackage{tabularx} 

\newcommand{\ba}{\begin{align}} 
\newcommand{\ea}{\end{align}}

\def\be#1\ee{\begin{align}#1\end{align}}
\def\bee#1\eee{\begin{align*}#1\end{align*}}

\def\E{\mathbb{E}}
\def\P{\mathbb{P}}

\def\bmx{\begin{pmatrix}}
\def\emx{\end{pmatrix}}

\theoremstyle{plain}
\newtheorem{theorem}{Theorem}
\newtheorem{lemma}[theorem]{Lemma}
\newtheorem{corollary}[theorem]{Corollary}
\theoremstyle{remark}

\theoremstyle{definition}
\newtheorem{definition}{Definition}

\newcommand{\cal}{\mathcal}

\newcommand{\mainfigwidth}{0.95\columnwidth}

\renewcommand{\vec}[1]{\ensuremath{\boldsymbol{#1}}} %

\newcommand{\coloneq}{\mathrel{\mathop:}=}
\newcommand{\eqcolon}{=\mathrel{\mathop:}}

\newcommand{\one}{\scalebox{1.05}{$\mathbf{1}$}}
\newcommand{\reals}{\ensuremath{\mathbb{R}}}

\newcommand{\trans}{^\intercal} %

\newcommand{\stoch}{\mathcal{M}} %
\newcommand{\MRPs}{\mathcal{R}}
\renewcommand{\S}{\mathcal{S}}
\newcommand{\tN}{\mathcal{N}}
\renewcommand{\P}{P}
\newcommand{\MCs}{\mathcal{P}}
\newcommand{\A}{\mathcal{A}}

\newcommand{\Obs}{\mathcal{O}}

\newcommand{\given}{\mid}
\DeclareMathOperator{\card}{card}

\edef\endfrontmatter{%
  \unexpanded\expandafter{\endfrontmatter}%
  \noexpand\endNoHyper %
}

\usepackage[colorlinks,allcolors=MidnightBlue,hypertexnames=true]{hyperref}

\begin{document}

\begin{frontmatter}
\title{Reinforcement Learning with Algorithms from Probabilistic Structure Estimation}
\author[JPESZ]{Jonathan~P.~Epperlein}
\author[RO]{Roman~Overko}
\author[JPESZ]{Sergiy~Zhuk}
\author[CK]{Christopher~King}
\author[DB]{Djallel Bouneffouf}
\author[ACRS]{Andrew Cullen}
\author[ACRS,CA]{Robert Shorten}
\address[JPESZ]{IBM Research Europe, Dublin, D15HN66, Ireland}
\address[RO]{Department of Electrical and Electronic Engineering at University College Dublin, Dublin, D04 R7R0, Ireland}
\address[CK]{Department of Mathematics at Northeastern University, Boston, MA 02115, USA}
\address[DB]{IBM Research Thomas J. Watson Research Centre, Yorktown Heights, NY 10598, USA}
\address[ACRS]{Dyson School of Design Engineering at Imperial College London, SW7 2AZ, UK} %
\address[CA]{Corresponding Author}


\begin{abstract}
Reinforcement learning (RL) algorithms aim to learn optimal decisions in unknown environments through experience of taking actions and observing the rewards gained. In some cases, the environment is not influenced by the actions of the RL agent, in which case the problem can be modeled as a contextual multi-armed bandit and lightweight \emph{myopic} algorithms can be employed. On the other hand, when the RL agent's actions affect the environment, the problem must be modeled as a Markov decision process and more complex RL algorithms are required which take the future effects of actions into account. Moreover, in practice, it is often unknown from the outset whether or not the agent's actions will impact the environment and it is therefore not possible to determine which RL algorithm is most fitting. In this work, we propose to avoid this difficult decision entirely and incorporate a choice mechanism into our RL framework. Rather than assuming a specific problem structure, we use a probabilistic structure estimation procedure based on a likelihood-ratio (LR) test to make a more informed selection of learning algorithm. We derive a sufficient condition under which myopic policies are optimal, present an LR test for this condition, and derive a bound on the regret of our framework. We provide examples of real-world scenarios where our framework is needed and provide extensive simulations to validate our approach. \end{abstract}
\end{frontmatter}

\section{Introduction}\label{sec: intro}
\let\svthefootnote\thefootnote
\let\thefootnote\relax\footnotetext{This work was partially supported by SFI grant 16/IA/4610.\\
Andrew Cullen is supported by an IBM PhD Fellowship.\\
This is a preprint of~\cite{automatica}, extended by including supplementary material in the appendix%
}
\let\thefootnote\svthefootnote
Markov decision processes (MDPs), in particular in the guise of reinforcement learning (RL) algorithms, are rapidly becoming a crucial tool in developing solutions for addressing new large-scale control and estimation problems that arise in the context of Smart Cities. Examples of their use are readily found in the literature ranging from crowdsourcing applications in the realm of smart mobility \cite{overko2020spatial} to problems arising in air traffic control \cite{kochenderfer2015decision}. More recently, RL agents have successfully achieved human-level control in interactions with other machines or with humans \cite{mnih2015human}. In the Smart Cities setting, such scenarios are both challenging and ubiquitous, giving rise to important problems in which agents must interact with highly uncertain environments. Typical examples of such problems arise when a number of independent agents compete to provide a service for some monetary remuneration. These agents, their state, and the manner in which they interact with each other, constitute an unknown environment. The control task is to develop an understanding of this environment, and based on this, to allocate demand among these agents in a manner that is optimal. RL algorithms are known to be effective in solving general problems of this nature~\cite{huang2011reinforcement}. However, practical applications give rise to new sets of challenges which make straightforward application of RL algorithms difficult. In particular, prior knowledge of the environment is often unavailable, and because it is often characterized by agents that have no incentive to reveal their true dynamics, identification of the environment structure can be very difficult. An example of such a {\em broker-supplier} relationship might arise in the case of a platform which allocates cars among a set of competing parking lots (or electric vehicle charging stations)~\cite{overko2021spatial}, or indeed an entity like {\em Amazon}  allocating product demand across a set of suppliers. This latter example in particular inspires the broker problem referred to throughout this work. It is precisely applications of this nature that are of interest to us.  

Although RL algorithms have proven extremely successful in many applications, some fundamental questions remain unresolved. One such question pertains to a choice facing the designer of any RL agent; namely, that of the \emph{learning algorithm} to be employed. This, along with many of the basic outstanding questions in RL, has strong parallels to those that have arisen in model predictive control, adaptive control, and even in classical control. {Other instances of problems of this kind from the control literature include selection of an appropriate controller/plant model, for example, multiple model adaptive control \cite{narendra1997adaptive}, and other structure identification methods in the identification literature. Such problems are closely related in spirit to the problem of RL algorithm selection.} Despite these similarities and the general links between control theory and RL (see the excellent recent survey \cite{recht2019tour}), solving these problems in an RL setting brings new difficulties requiring the solution of sets of bespoke challenges. This manuscript seeks to address one such instance, namely it considers cases in which the choice of an appropriate learning algorithm is not obvious or intuitive, and presents a tool which allows this choice to be made automatically. RL, at its core, deals with environments that are unknown, but the designer typically must possess some degree of knowledge about the environment to accurately model the problem. Specifically, RL is concerned with sequential decision making problems modeled as MDPs, as we shall discuss in detail in the next section. The designer must model the environment's state, and the actions available to the agent, and then must choose an appropriate learning algorithm depending on whether or not the actions of the agent affect the environment. In particular, two broad classes of algorithm can be considered: those based on contextual multi-armed bandits (CMABs)~\cite{abbasi2011improved}, which are appropriate for problems in which the agent's actions do not impact the environment; or those based on a more general MDP model (e.g., \emph{Q-Learning}~\cite{watkins1992q}), in which the agent's actions do impact the environment.  

In many examples of classic RL settings, the choice of algorithm is quite obvious. For example, if the agent is tasked with navigating through a physical environment to find some goal state, it is intuitive to use an MDP-based algorithm, as the agents actions (movement) clearly change the state (position of the agent in the environment)~\cite{overko2020spatial},~\cite{khalid2020reinforcement}. Many other problems, however, do not exhibit such a clear relationship between the actions of the agent and the state of its environment, and these problems are the subject of this paper. Specifically, we will present a mechanism which continuously tests a hypothesis about the relationship between the actions of an RL agent and its environment and selects the appropriate learning algorithm automatically. We call this algorithm selection mechanism an \emph{orchestrator}. We refer to our framework as RLAPSE---Reinforcement Learning with Algorithms from Probabilistic Structure Estimation. RLAPSE consists of the orchestrator and two RL algorithms between which it can choose, one for each potential underlying environment structure.  

\begin{figure}[h]
    \centering
    \includegraphics[width=\columnwidth]{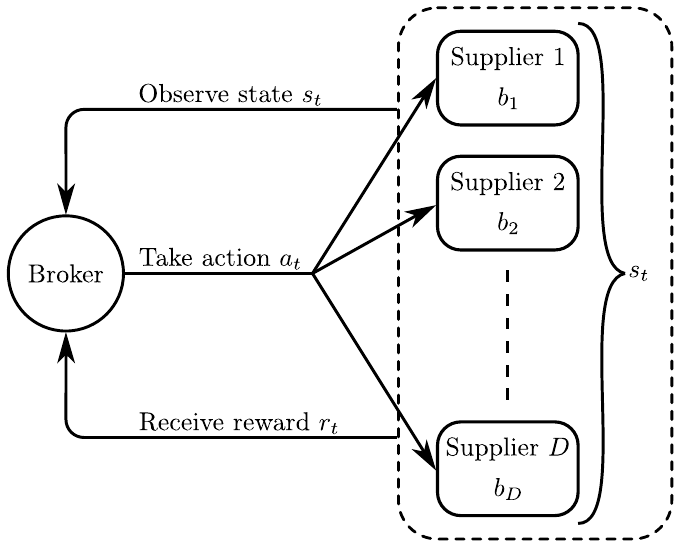}
    \caption{The broker observes the state $s_t$ of the suppliers (prices offered) at time $t$ and selects action $a_t$ (a supplier). The broker then receives reward $r_t$ which depends on $s_t$ and $a_t$. The state of the suppliers then transitions to $s_{t+1}$.}
    \label{fig: broker}
\end{figure}

Consider an illustrative example of a broker choosing among a set of suppliers from whom to buy a commodity, as depicted in Figure \ref{fig: broker}, which represents an archetype for the broker-supplier relationship mentioned above. The state in this RL problem represents a ``snapshot'' of the prices offered by each supplier for a unit of the commodity, and actions correspond to choosing a particular supplier to buy from. The goal of the broker here is to maximize the reward gained by buying the best quality commodities at the lowest possible price from the suppliers. At first glance, a CMAB may seem to offer an ideal formulation of this problem because it seems unlikely that the action (choosing a supplier) should affect the future state (prices offered). However, the supplier could be tailoring its prices to each of its customers based on that customer's past purchasing behavior, with the goal of maximizing the supplier's own profit. In this case, the actions (choosing a supplier) of the RL agent (broker) would indeed affect the state (prices offered). It is therefore likely that the solution found by a CMAB-based algorithm would not be optimal as it would seek only to maximize the immediate reward rather than learning a best response to the unknown dynamics of the suppliers as an MDP-based algorithm could allow. It is impossible to predict how new environments will respond to the actions of the broker when other autonomous actors (suppliers) are involved. It is especially difficult to identify such effects when the algorithms employed by these actors that make up the environment (suppliers) are intentionally kept secret, perhaps to protect intellectual property or even to prevent exploitation by other agents (brokers). We elaborate further on this example in Section \ref{sec: eval}.  

\subsection{Paper structure}
In Section \ref{sec: primer}, we present a primer on RL and the processes which underlie RL problems, namely MDPs. In Section \ref{sec: rlapse}, we introduce the RLAPSE framework, which consists of a suite of RL algorithms and a mechanism for choosing the most appropriate one based on observations from the environment. In Section \ref{sec: results}, we present our main theoretical results which begin by establishing a sufficient condition under which a CMAB-based learning algorithm finds an optimal policy and we subsequently find a regret bound for the RLAPSE framework. In Section \ref{sec: eval}, we evaluate the efficacy of RLAPSE through simulation of the system employed on an example environment based on the broker-supplier setting. Additionally, we perform simulations on a set of randomly generated MDPs. Finally, in Section \ref{sec: conc}, we summarize our contributions and conclude with a statement on the broader impact of this work.
 
\section{Primer on reinforcement learning}\label{sec: primer}
RL algorithms aim to learn how to make optimal decisions in an unknown environment entirely through experience of interacting with that environment \cite{sutton2018reinforcement}. Specifically, RL applies to sequential decision problems modeled as MDPs, in which the agent observes the environment state $s_t$ at time $t$ and takes action $a_t$. The agent then receives a reward $r_t$ from the environment. As the name of the process suggests, the Markov assumption holds, namely the next observed state $s_{t+1}$ depends only on $s_t$ and $a_t$, and not on any previous states or actions. The probability distribution of the first state $S_0$ is denoted by $\varpi$, the state transition probabilities are given by $P(S_{t+1}=j | S_{t}=i, A_t=a)=p_{ij}(a)$, and the reward matrix is given by $r(S_t = s, A_t = a)$. Consider an MDP with $N$ discrete states and $A$ actions. The transition dynamics can be thought of as a 3-dimensional $N\times N\times A$ tensor of transition probabilities $P(:)$. The MDP is then fully parametrized by a tuple $(P(:), \varpi, r)$. Each action $a$ gives rise to an $N\times N$ transition matrix $P(a)$, which we refer to as a \emph{page} of the tensor $P(:)$, and whose $(i,j)$ element corresponds to the probability that the state transitions from $i$ to $j$ when action $a$ is taken. Agents acting in MDP environments employ a \emph{policy} $\pi$ to decide which action to take given the current state of the environment. The policy maps the state to a probability distribution over the actions, where $\pi(a|s)$ denotes the probability of choosing action $a$ in state $s$.

RL is concerned with learning how to act optimally in an MDP environment for which the parameters, $(P(:), \varpi, r)$, are not known. Two classes of MDP environments are of interest in this work:

\begin{itemize}
\item \emph{uncontrolled MDPs}---state transition probabilities do not depend on actions, i.e., all pages of $P(:)$ are equal; and
\item \emph{controlled MDPs}---state transition probabilities depend on actions, i.e., the pages of $P(:)$ are not all equal.
\end{itemize}

Uncontrolled MDPs, being a special case of controlled MDPs, require less complex algorithms to solve. Next, we discuss some common RL algorithms for solving each of these classes of MDP.

\subsection{RL for uncontrolled MDPs}
The state of the art in RL for uncontrolled MDPs can be captured with discussion of the contextual multi-armed bandit (CMAB) problem \cite{li2010contextual}. The CMAB augments the well-studied multi-armed bandit problem which models a slot machine with multiple arms and an unknown reward distribution over those arms. A multi-armed bandit algorithm must balance exploration with exploitation to build a sufficiently accurate model of all arms without wasting opportunities to gain reward from the best arms. In a CMAB environment, the agent is additionally presented with a state, or context, which modulates the reward received for choosing each arm, so the agent must also take this context into account when choosing an arm.  

CMAB problems are uncontrolled MDPs, that is, the state at the next time step is independent of the action chosen at the current time step, so choices made by the agent do not have any role in determining future states. Additionally, it is typically assumed that the next state is independent of the current state in CMABs. In other words, all pages $P(a)$ of the transition probability tensor are identical to one another and the rows of $P(a)$ are all equal, meaning that the next state is simply drawn from an i.i.d.\ distribution given by any row of $P(a)$. The implication of this simplified structure of the MDPs underlying CMABs is that RL algorithms for CMABs do not need to take future states into account, because the agent's actions have no bearing on this. We refer to such short-sighted RL algorithms as \emph{myopic} \cite{sutton2018reinforcement}. Compared to more general RL algorithms which must take future state trajectories into account, myopic algorithms save hugely on complexity.   

Myopic RL algorithms for CMAB environments have been successfully deployed in a variety of real scenarios, such as web-page recommendation \cite{li2010contextual} in which the agent learns to choose which news article to show the current customer in order to maximize revenue. In this example, the context is provided by certain information about the customer's profile, the actions correspond to choosing a news article to show from a suite of options, and a reward is gained in the form of revenue when a customer clicks the article.

\subsection{RL for controlled MDPs}
In the more general case of RL, it is assumed that the actions of an agent do affect its environment, and specifically that they modulate the state transition probabilities. In other words, the environment is typically assumed to be a controlled MDP. The consequence of this is that an RL agent acting in this more general environment than the CMAB setting must consider not only the immediate reward received for its actions, but also potential for future rewards. This consideration is captured by the return function which is a discounted sum of the rewards received each time step in the future:
\begin{multline}\label{eq: return}
    G_t = r(S_t, A_{t}) + \gamma r(S_{t+1},A_{t+1}) \\
    + \gamma^2 r(S_{t+2},A_{t+2})  + \cdots. 
\end{multline}
Most notable here is the discount factor $\gamma \in [0,1)$ which determines how much importance is placed on future rewards in calculating the return. Employing a return with $\gamma=0$ yields myopic policies, comparable to those produced by a CMAB-based algorithm as discussed above.  

The most widely used RL algorithms are concerned with estimating the \emph{action-value} functions associated with policies:
\begin{equation}
q_\pi(s,a) = \mathbb{E}_\pi [G_t \given S_t=s, A_t =a].
\end{equation}
In other words, $q_\pi(s,a)$ is the expected return at time $t$ in state $s$ taking action $a$ and subsequently following policy $\pi$. Perhaps the most popular action-value based RL algorithms are based on temporal difference methods, for example Q-learning~\cite{watkins1992q}, and can estimate action-values in an incremental and online fashion. Consider the update used in Q-learning, carried out after taking action $a_t$ in state $s_t$ and observing the next state $s_{t+1}$ and the reward $r_{t}$:
\begin{multline}\label{eq: qupdate}
Q(s_t, a_t) \gets Q(s_t, a_t) + \alpha[r_{t} \\
+ \gamma \max_{a}Q(s_{t+1},a) - Q(s_t,a_t)].
\end{multline}
Here, the Q-value is updated using a newly observed reward $r_{t}$ along with maximum of the existing estimates for $Q_(s_{t+1},a)$ and with learning rate $\alpha$; $\alpha$ has to be designed carefully, see \cite{even2003learning} for the approach used in the evaluations in Section~\ref{sec: eval}. Using existing estimates in this fashion for updates is known as \emph{bootstrapping}, and in the case of Q-learning, it can be proven that the Q-values converge to $q^*$, the action-value function of the optimal policy. Q-learning is an off-policy RL algorithm since it estimates the optimal action-value independently from the policy followed. The actual policy followed should allow for some exploration of the state-action space, for example, an $\varepsilon$-greedy policy chooses the action $a_t$ which maximizes $Q(s_t, a_t)$ with probability $1-\varepsilon$, and selects an action uniformly at random with probability $\varepsilon$. SARSA is the name given to a similar on-policy learning algorithm, in which the estimate $Q(s_{t+1}, a_{t+1})$ associated with the action taken at the next step  is used to bootstrap the update, rather than the maximum over the actions as in \eqref{eq: qupdate}.  

As we shall see in Section \ref{sec: results}, an important metric for the performance of an RL algorithm is regret. The regret of an algorithm $\A$ accumulated during $T$ time steps is 
\begin{equation}\label{eq: regret}
R(T) \coloneq \sum ^{T}_{t=1} r(S_t,A^*_t) - \sum^{T}_{t=1} r(S_t,A_t),
\end{equation}
where $A^*_t$ is the optimal action at time $t$, and $A_t$ is the one chosen by $\A$. Clearly, this metric reflects whether or not an algorithm finds an optimal policy, and also how quickly that policy is found. When an RL algorithm designed for a controlled MDP is employed in an uncontrolled MDP, this can lead to slow convergence to the optimal policy, and hence a high regret. One of the main results presented in this work is a regret bound for the RLAPSE framework.  

Notable advances in this area since the development of Q-learning have included the use of function approximation in algorithms such as in Deep Q-learning, which allows agents to learn in extremely large state spaces. Additionally, methods which work directly with policies rather than action-values have been proposed. These methods allow agents to naturally deal with large and continuous state-action spaces. We do not provide any further overview of such techniques as the focus of this work is on \emph{tabular} methods, in which the size of the state-action space permits estimates for all state-action pairs to be stored in a table. We defer the issue of approximated state representations and larger or continuous state-action spaces to future work.
 
\section{Reinforcement Learning with Algorithms from Probabilistic Structure Estimation}\label{sec: rlapse}
In the preceding sections, we identified two distinct classes of MDP, namely controlled and uncontrolled. Moreover, we highlighted the implications of this distinction when it comes to selecting an appropriate RL algorithm: myopic algorithms, such as those designed for CMABs, can learn to act optimally in uncontrolled MDP environments, but in the case of controlled MDPs, more complex RL algorithms are required, which take future rewards into account. Although model uncertainty is at the core of any RL problem, it is often taken for granted that an algorithm designer will at least be certain of whether the underlying MDP is controlled or uncontrolled. Furthermore, making the incorrect assumption can have detrimental effects on the performance of the designed RL agent. The RLAPSE framework takes this choice out of the hands of the designer and makes it automatically with a likelihood-ratio (LR) orchestrator.  

In short, the RLAPSE framework consists of three components:
\begin{enumerate}
\item a myopic RL algorithm $\A_0$, which is suitable for uncontrolled MDPs; 
\item a full RL algorithm $\A_1$, which considers the agent's future potential to earn rewards to some horizon, and is hence capable of learning in controlled MDPs; and
\item the LR orchestrator, which detects whether the MDP environment is uncontrolled or controlled and deploys algorithm $\A_0$ or $\A_1$, respectively.
\end{enumerate}

The following section will present the main results which include the design of the LR orchestrator.

\subsection{Related work}
To the best of our knowledge, our work is the first to consider the dichotomy between controlled and uncontrolled MDPs, but prior research has dealt with the related problem of finding the optimal horizon to be used for calculating returns for a given RL algorithm. In~\cite{jiang2015dependence}, the authors discuss choosing a shorter planning horizon to achieve faster convergence of RL algorithms. \cite{xu2018meta} presents a technique for learning optimal ``meta-parameters'' of return functions. Meta-parameters of the return function can include, for example, the discount factor $\gamma$ (see Equation \eqref{eq: return}), which determines how much weight is given to future rewards when estimating action-value functions. The optimal meta-parameters are highly dependent on the agent's environment and their choice is difficult for even the most skilled designers. In~\cite{xu2018meta}, the authors deal with the choice of horizon within a given RL algorithm, while RLAPSE automates the choice of algorithm based on observations of state transitions. The approach of~\cite{xu2018meta} and our own can hence be thought of as complementary.   

A topic related to our own is also discussed in~\cite{zanette2019UBEVS}. Here, a specific  sample-efficient episodic MDP algorithm that can ``adapt'' to the environment in the sense that it achieves a better regret bound for CMABs is presented. In contrast, RLAPSE can be combined with arbitrary RL algorithms and with high probability it picks the appropriate one explicitly, and thus asymptotically achieves its regret. Additionally, we consider the more general concept of uncontrolled MDP instead of CMABs.  

Recent work on industrial recommendation systems has also identified algorithm selection as an important issue in practical applications. The authors of \cite{Mitra2020AutomatedMT} also consider the problem of choosing between RL and CMAB algorithms, although their work does not consider the underlying MDP whatsoever, and no theoretical results are provided. Their approach is based on trying each type of algorithm and observing the performance to choose the most appropriate.  

Our aim is to design an algorithm which interacts with an unknown environment and determines whether it is an uncontrolled MDP, in which case a myopic policy is optimal, or not. The steps taken to design and analyze this algorithm are as follows: 
\begin{enumerate}[A),align=left,nosep]
    \item  \textit{optimality condition for myopic policies:} establish a sufficient condition under which a myopic policy is optimal;
    \item \textit{LR orchestrator:} design a computational procedure which tests the condition of A) during the interactions with an unknown environment.
    \item \textit{regret bound for RLAPSE:} We then derive a bound on the probability of error and resulting regret of the procedure of B). 
\end{enumerate} 
\section{Main results}\label{sec: results}
Throughout, we assume that the environment is an MDP with deterministic rewards.

\subsection*{Preliminaries}
Let $\stoch^{1\times N}$ denote all stochastic row vectors, or equivalently, all discrete distributions over the numbers $[N]\coloneq\{1,2,\dotsc,N\}$;  $\stoch^N \coloneq \stoch^{N\times N}$ denotes all row-stochastic $N\times N$ matrices; finally, $\stoch^{N\times N\times A}$ refers to all 3-dimensional tensors whose pages $P(a)$ are in $\stoch^N$, or in other words, $\sum_{j=1}^N [P(a)]_{ij}=1$ $\forall j,a$. A vector of $N$ ones will be denoted by $\one_N$ or just $\one$ if the dimension is clear from context. The notation $\E_X\{f(X)\}$ and $\E_p\{f(X)\}$ will denote the expected value of the random variable $f(X)$ with respect to the distribution of $X$ and $p$, respectively, where the subscript is optional. The remainder of this section closely follows~\cite{puterman1994markov}.  

A \textbf{Markov chain (MC)} on the state space $\S$ with $|\S| = N$ is  parametrized by a tuple $(P,\varpi)$, where $\varpi\in\stoch^{1\times N}$ is the probability distribution of the initial state $s_0$, and the matrix $P\in\stoch^{N}$ with entries $p_{ij}$ is its \emph{transition probability matrix}: \(
P(S_t = j \given S_{t-1}=i) = p_{ij}.
\)

A \textbf{Markov reward process (MRP)} is represented as a tuple $(P, \varpi, r)$ and adds a reward function $r: \S\mapsto\reals$ to the MC $(P,\varpi)$. At each time $t$, the reward $r(S_t)$ is collected; in general, rewards are stochastic themselves, but here we consider the simpler case of a deterministic reward function.

A \textbf{Markov decision process (MDP)} adds to an MRP a set of actions $\A = [A]$ which modulate the transition probabilities and rewards, i.e., at each time step $t$, an action $A_t\in\A$ is chosen, and the transition probabilities and reward are now given by
$P(S_t = j \given S_{t-1}=i, A_t = a) = p_{ij}(a)$,  $r(S_t = s, A_t=a)$.
The reader can refer to Section \ref{sec: primer} for further terminology related to MDPs including definitions of controlled and uncontrolled MDPs which are particularly relevant to this work.  

\emph{Policies}, as discussed in Section \ref{sec: primer}, are used to choose actions. We consider two types of policies: a \emph{Markov randomized (MR)} policy is a mapping $\pi: \S \mapsto \stoch^{1\times A}$, so that if the state at time $t$ is $s_t$, then the action at time $t$ is chosen according to $P(A_{t} = a \given S_t = s) = \pi_a(s)$; a \emph{Markov deterministic (MD)} policy is a mapping $\pi: \S \mapsto \A$, so that we have $a_{t} = \pi(s_t)$. Recall from Section \ref{sec: primer} that a myopic policy, which is denoted by $\pi^C$, is one that in each state $s$ selects an action that provides maximal immediate reward, i.e., $\pi^C(s) = \text{argmax}_a r(s,a)$.

Since MD policies are special cases of MR policies, we can describe any policy we consider here by a matrix $\Pi\in\stoch^{N\times A}$, whose $i$-th row is the stochastic vector $\pi(i)$.  

It is instructive to think of policies as closing the loop between actions and states: an MDP $(P(:), \varpi, r)$ is a non-autonomous system with inputs in the form of actions, whereas once a policy is specified, the combined system of MDP and policy is autonomous. If the policy is MR and given by $\Pi$, then this autonomous system is a MRP $(P_\pi, \varpi, r_\pi)$, with the transition matrix and rewards vector given by
\begin{align*}
    (P_\pi)_{ij} %
    &\coloneq \sum\nolimits_{a\in A} \pi_a(i) P(j \given i, a),\\
    (r_\pi)_i %
    &\coloneq \sum\nolimits_a r(i,a) \pi_a(i).
\end{align*}
Denote by $\MRPs(P(:), \varpi, r)$ the set of all MRPs that can be generated from the MDP $M = (P(:), \varpi, r)$ by a MR policy, and specifically denote by $\MCs(M)$ the (convex) set of their transition matrices $P_\pi$, i.e.,
\begin{multline}\label{eq:MCs}
    \MCs(M) \coloneq 
    \Bigl\{ P_\pi \given 
        (P_\pi)_{ij} = \sum\nolimits_{a\in A} \pi_a(i) P(j \given i, a),\\ 
     \pi:\S\mapsto\stoch^{1\times A}\;\forall\, i,j\in \S\Bigr\}.
\end{multline}

We recall a few necessary facts from the theory of MCs (see \cite{puterman1994markov} for details).
\begin{definition}\label{def:MCs}
    A subset $C\subseteq [N]$ of states of an MC $(P,\varpi)$ is 
    \textbf{closed and irreducible} if for every pair of states $i,j\in C$ %
    there is an $n_{ij}<\infty$ such that $\left(P^{n_{ij}}\right)_{ij}>0$, and if for every $k\not\in C$, $P_{ik}=0$.
    
    A MC is a \textbf{unichain} if its state space consists of only one closed and irreducible subset and one (possibly empty) subset of transient states.\footnote{%
        A state is recurrent, if it is visited infinitely often as $t\rightarrow\infty$. States that are not recurrent are transient.}
\end{definition}
For every unichain MC $(P,\varpi)$, there exists a unique non-negative left-eigenvector $w\trans\in\stoch^{1\times N}$ of $P$, called the \textbf{Perron vector}, corresponding to the eigenvalue 1, i.e., there is a unique solution $w\trans$ for
\begin{align*}
    w\trans P &= w\trans, & w\trans \one &= 1, & w_i&\geq 0 & \forall i \in [N].
\end{align*}
\begin{definition}\label{def:MDP_unichain}
    An MDP $(P(:), \varpi, r)$ is \textbf{unichain}, if \emph{all} matrices $Q\in\MCs(P)$ correspond to unichains.
\end{definition}
Lastly, coefficients of ergodicity can be defined in more generality, but we only need them for $\stoch^N$. They can be used to estimate convergence rates, eigenvalue locations, and most importantly for us, the sensitivity of Perron vectors to perturbations.
\begin{definition} \cite{Ipsen2011} \label{def:tau1}
    The \textbf{(1-norm) coefficient of ergodicity} of $P\in\stoch^N$ is 
    \begin{equation}\label{eq:tau1proj}
        \tau_1(P) \coloneq \max_{\substack{\|z\|_1=1 \\ z\trans\one=0}} \|P\trans z\|_1 
        = \frac{1}{2} \max_{i,j} \sum_k | P_{ik} - P_{jk} |.
    \end{equation}
\end{definition}
\begin{lemma} \cite{seneta2006non} \label{lem:scramble}
    Let $P\in\stoch^N$. Then,
    \begin{itemize}
        \item $\tau_1(P) = 0$ iff $P$ is rank-1, i.e.\ iff $P  = \one p\trans$, hence every row of $P$ equals $p\trans$;
        \item $\tau_1(P)<1$ iff no two rows are orthogonal, or equivalently, iff any two rows have at least one positive element in the same column. In this case, $P$ is called \textbf{scrambling}.
    \end{itemize}
\end{lemma}

\subsection{Optimality condition for myopic policies}
We begin by showing that the environment being an uncontrolled MDP is sufficient for a myopic policy to be optimal. We do this by answering a slightly more general question, namely:  
\emph{given full knowledge of the parameters of an MDP, can we compute the decrease in average reward if a myopic policy is sought?}
Theorem~\ref{thm:bound} provides an upper bound on how much worse myopic policies perform than optimal ones; Corollary~\ref{cor:myopic} identifies uncontrolled MDPs as the environments for which this bound equals 0.
For ease of analysis, we consider the expected average reward as the criterion to be optimized, which corresponds to a discount factor (or ``evaluation horizon'') $\gamma=1$.%
\begin{definition} The \textbf{average reward} or \textbf{gain} of a policy $\pi$ is 
    \begin{equation}\label{eq:gammapi}
        \Gamma_\pi(s) \coloneq \lim_{n\rightarrow\infty} \frac{1}{n} \E_\pi \left\{ \sum\nolimits_{t=1}^{n} r(S_t, A_t) \right\}.
    \end{equation}
\end{definition}
This limit need not exist in the general case. However, for \emph{unichain} MDPs under Markov randomized policies it does exist, and it is \emph{independent of the initial state} $s$. This constant gain is denoted by $\Gamma_\pi(s)\equiv g_\pi$ for unichain MDPs. See also Lemma~\ref{lem:gain} in the Supplementary Material and e.g.,~\cite[Ch.~8]{puterman1994markov} for more details.  

It then follows that for a unichain MDP $M=(P(:), \varpi, r)$, there is a Markov randomized policy $\pi^*$ which achieves the optimal average reward
$g^* \coloneq g_{\pi^*} = w^* r_{\pi^*},$
where $w^*\coloneq w_{\pi^*}\trans$ is the Perron vector of $P_{\pi^*}$.  

Define the matrices $\overline{P}_{ij} = \max_a P(j \given i, a)$, $Y \in \MCs(M)$ and $\underline{P}_{ij} = \min_a P(j \given i, a)$, %
and 
\begin{equation}\label{eq:Einfty}
    \rho(M,Y) \coloneq 
    \max\limits_i \sum\limits_j \frac{\overline{P}_{ij}-\underline{P}_{ij}}{2}  \!+\! \left| Y_{ij} - \frac{\overline{P}_{ij}+\underline{P}_{ij}}{2}\right|.
\end{equation}
Then, $\rho(M,Y)$ is the ``radius'' of the set $\MCs(M)$ centered around $Y$ in the sense that
\begin{equation} \label{eq:PMDPeps}
    \MCs(M) \subseteq \left\{ Y+E \given \|E\|_\infty \leq \rho(M, Y) \right\}.
\end{equation}
Geometrically, this representation shows the set of all MCs that can  be generated from the MDP $M$ as being contained in a ball of radius $\rho(M, Y)$ centered at $Y$. %
\begin{theorem}\label{thm:bound}
    Let $P_C=P_{\pi^C}$ be a scrambling transition matrix of a unichain MC, and let $g^C$ and $g^*$ be the gain of the myopic policy $\pi^C$ and the optimal policy $\pi^*$, respectively. 
    Then,
    the optimal policy outperforms the myopic policy by no more than the following bound:
    \begin{equation}\label{eq:gapinf_tau1}
        \frac{g^* - g^C}{\overline{r}} \leq \frac{1}{1-\tau_1(P_C)} \rho(M, P_C),
    \end{equation}
    where $\rho(M,P_C)$ is defined in \eqref{eq:Einfty}, $\overline{r} = \max_{s,a} r(s,a)$ is the maximal available reward, and $\tau_1(\cdot)$ is defined in~\eqref{eq:tau1proj}. %
    
    See Theorem~\ref{theorem: gap} in Section~\ref{ssec:sup:perfgap} of the appendix for details and proof.
\end{theorem}
Intuitively, the two factors in the upper bound in~\eqref{eq:gapinf_tau1} quantify the two aspects in which a controlled MDP can differ from a CMAB. The first term is minimized when $\tau_1(P_C)=0$, meaning that $P_C$ is a rank-one matrix, i.e., one in which the \emph{current state} has no influence on the next state. The second term, $\rho(M,P_C)$, measures how much influence the \emph{current action} can have on the next state and it equals 0 iff the MDP is uncontrolled.
This last fact can be stated formally as follows.%
\begin{corollary}\label{cor:myopic}
    If the environment is an uncontrolled MDP, then $g^C=g^*$. In other words, the myopic policy is also optimal.
\end{corollary} 

\subsection{Likelihood-ratio orchestrator}
The main tool in designing the orchestrator is the \textbf{LR test}~\cite{kalbfleisch1985probability,wilks1938}, which is used in a classical context to test \emph{nested} model structures (when one model is a special case of the other); in our case, an uncontrolled
MDP (model $M_0$) is a special case of a controlled MDP (model $M_1$), hence the model structures are nested.  

For a given observation sequence $\Obs$, we compute the maximum-likelihood estimates of the parameters of the two models $M_0$ and $M_1$, with $M_0$ being nested into $M_1$. Denote the corresponding likelihoods by $\ell_0 \coloneq P(\Obs \mid \hat\theta_0)$ and $\ell_1 \coloneq P(\Obs \mid \hat\theta_1)$, so that the ratio of likelihoods $\lambda \coloneq \ell_0/\ell_1$ is always in $[0,1]$ since $M_1$ is more general than $M_0$.  

The test statistic used is
$
L \coloneq -2 \ln{\lambda}
$
and Wilks' Theorem~\cite{wilks1938} states that \emph{if $M_0$ is the correct model structure underlying $\Obs$}, then, as the number of samples goes to $\infty$,  $L$ asymptotically follows a $\chi^2_k$ distribution, where $k$ is the difference in degrees of freedom between $M_1$ and $M_0$. Denote by $F$ the CDF of a $\chi^2_k$-distributed random variable, i.e.,
\(
X\sim \chi^2_k \quad \Rightarrow F(x) = P( X\leq x  ).
\)
The LR test with significance level $\alpha$  then rejects the hypothesis that $M_0$ is the correct model structure if the probability of obtaining a value at least as large as $L$ under the assumption that $M_0$ is the correct structure is less than $\alpha$, i.e., if $1 - F(L) \leq \alpha$.  

As per Corollary~\ref{cor:myopic}, a myopic policy is optimal if the next state does not depend on the current action. Hence, we say that under hypothesis $H_0$, $M_0$ is the correct model, i.e., all pages of $P(:)$ are equal, whereas under the alternative hypothesis $H_1$, they are not.
\begin{align*}
    H_0{:}\,\,P(S_t =s \mid s_{t-1}, a_{t-1}) &= P(S_t = s \mid s_{t-1}),\\ %
    H_1{:}\,\,P(S_t =s \mid s_{t-1}, a_{t-1}) &= P(S_t =s \mid s_{t-1},a_{t-1}). %
\end{align*}
We may assume that the initial probabilities $P(S_0=s)$ are known, e.g., drawn from a uniform distribution $P(S_0=s)=1/N$. This is reasonable since we have no means of estimating them and they cancel in the likelihood ratio, $\lambda$, defined above anyway.  

The model $M_0$ has $N(N-1)$ parameters, 
whereas $M_1$ has $AN(N-1)$ parameters ($N$ and $AN$ parameters are fixed by
the stochasticity constraints in $M_0$ and $M_1$, respectively), and hence the difference in degrees of freedom is $k=N(A-1)(N-1)$.  

Now assume we have observations 
\[
\Obs = \left((s_0, a_0,r_0), (s_1, a_1,r_1), \dotsc, (s_T, a_T,r_T) \right).
\]
Define the transition counts
\begin{equation}
\begin{split}
    m(s',s,a) &= \card\bigl\{ t \given \\
            & \phantom{=} s_t=s', (s_{t-1},a_{t-1})=(s,a)\bigr\}, \\
    n(s,a) &= \sum\nolimits_{s'=1}^N m(s',s,a),\\
    m'(s', s) &= \sum\nolimits_{a=1}^A m(s',s,a),\\
    n'(s) &= \sum\nolimits_{s'=1}^N m'(s',s),
\end{split}
\label{eq:counts}
\end{equation}
hence $m(s',s,a)$ equals the number of times state $s'$ has been observed as the next state when action $a$ has been chosen in state $s$. We then get the maximum likelihoods of $M_0$ and $M_1$, and the test statistic $L$ as
\begin{equation}
\begin{split}
    \ell_0 &= \frac{1}{N}\prod_{s'=1}^N \prod_{s=1}^N \left(\frac{m'(s',s)}{n'(s)}\right)^{m'(s',s)}\\
    \ell_1 &= \frac{1}{N} \prod\limits_{(s,a): n(s,a)\geq1} \prod_{s'=1}^N 
    \left( \frac{m(s',s,a)}{n(s,a)} \right)^{m(s',s,a)}\!\!\!\!\\
    L & =-2 \ln{\ell_0/\ell_1}.
\end{split}
\label{eq:L}
\end{equation}
The computations to obtain these expressions can be found in~\ref{ssec:sup:LR} of the Supplementary Material.
\begin{algorithm}[tb]
    \caption{RLAPSE with LR Orchestrator}
    \label{alg:LRorch}
    \begin{algorithmic}[1]
        \STATE {\bfseries Input:} $\alpha\in[0,1]$, algorithm $\A_0$ seeking myopic policies, algorithm $\A_1$ with longer horizon. Number of states $N$, number of actions $A$, minimum amount of data $T_0$.
        \STATE {\bfseries }\textbf{Init:} $i=m(s',s,a)=m(s,a)=m(s)=T=0$ $\forall s,s'\in[N], a\in[A]$, $\Obs=()$.
        \FOR{$t=0,\dotsc,\infty$}
        \STATE Receive $s_t$, choose action $a_t$ according to $\A_i$.
        \STATE Receive $r_t$, update $\Obs$ with $(s_t,a_t,r_t)$.
        \STATE %
        Update \emph{both} $\A_0$ and $\A_1$ with $(s_t,a_t,r_t)$.
        \STATE $T++$
        \STATE Update $m, m', n, n'$ as in \eqref{eq:counts} \COMMENT{if that has not happened in the course of $\A_i$ anyway}
        \IF{$T>T_0$}
        Compute $L$ and $F(L)$ as in \eqref{eq:L}
        \IIf{$F(L)\geq1-\alpha$}
        $i = 1$ \COMMENT{i.e., use $\A_1$}
        \IElse%
        $i = 0$ \COMMENT{i.e., use $\A_0$}
        \ENDIIf
        \ENDIF 
        \ENDFOR

    \end{algorithmic}
\end{algorithm}  
It is then straightforward to compare $F(L)$ to $1-\alpha$. Algorithm~\ref{alg:LRorch} outlines the full RLAPSE framework with LR orchestrator. In numerical experiments (Section~\ref{sec: eval}) specific choices of $\A_0$ and $\A_1$ are made.

A practical note concerning the applicability of the LR test to small sample sizes: since the $\chi^2_k$ distribution is only followed asymptotically (as the number $T$ of observations goes to $\infty$), it makes no sense to enable the orchestrator from the beginning, but rather let it collect a ``reasonable'' amount of data $T_0$ before performing the first test.

{
Note that Algorithm~\ref{alg:LRorch} requires storing of transition counts defined 
in~(\ref{eq:counts}). For large-scale MDPs, storing of these transition counts, especially $m(s',s,a)$, 
might lead to high space complexity of the algorithm. However, it is highly likely that some of the 
combinations of $(s',s,a)$ will never be observed in environments with large state-action spaces. 
Thus, it is natural to implement the transition counts stored by the RLAPSE algorithm as sparse matrices.
In fact, a sparse matrix is also an appropriate data structure to store Q-values in the Q-learning 
algorithm in order to alleviate high space complexity for large-scale problems. Therefore, we highly 
recommend implementing the transition counts~(\ref{eq:counts}) as sparse matrices.

It is also worth mentioning that, for large-scale MDPs, the RLAPSE algorithm might introduce additional
computational complexity via calculating the maximum likelihoods $\ell_0$ and $\ell_1$~(\ref{eq:L}) at each
time step after $T_0$. Moreover, direct implementation of equations~(\ref{eq:L}) might lead to 
floating-point arithmetic issues (e.g., arithmetic underflow) in case of extremely small values of 
$\ell_0$ and $\ell_1$. We therefore present an incremental version of updating the test statistic $L$ in 
Algorithm~\ref{alg:incrementalL} in order to mitigate extra computational complexity of RLAPSE for
large-scale problems. Additionally, Algorithm~\ref{alg:incrementalL} deals with log-likelihoods
$L_0 = \ln{\ell_0}$ and $L_1 = \ln{\ell_1}$ (rather than $\ell_0$ and $\ell_1$) to provide safe 
floating-point operations. 
Note that such an incremental approach requires computing $L_0$ and $L_1$ 
at each time step regardless whether the orchestrator is started or not. However, for large-scale 
MDPs, the incremental version significantly reduces computational complexity of the likelihoods. 

\begin{algorithm}[tb]
    \caption{Incremental computation of $L$}
    \label{alg:incrementalL}
    \begin{algorithmic}[1]
        \STATE {\bfseries }\textbf{Init:} $L_0 = L_1 = -\ln{N}$.

        \STATE {\bfseries }\textbf{Define:}
        \STATE {\bfseries }\textbf{\,\,\,\,\,\,function} $f_0(m', n', s)$
        \STATE \,\,\,\,\,\,\,\,\,\,\,\,$x(s') := m'(s', s),\,\,y := n'(s)\,\,\forall s' \in [N]$.
        \STATE {\bfseries }\textbf{\,\,\,\,\,\,\,\,\,\,\,\,return} $\sum_{s'=1}^{N}
            x(s')\left[\ln{x(s')} - \ln{y}\right]$
        \STATE {\bfseries }\textbf{\,\,\,\,\,\,end function}

        \STATE {\bfseries }\textbf{\,\,\,\,\,\,function} $f_1(m, n, s, a)$
        \STATE \,\,\,\,\,\,\,\,\,\,\,\,$x(s') := m(s', s, a),\,\,y := n(s, a)\,\,\forall s' \in [N]$.
        \STATE {\bfseries }\textbf{\,\,\,\,\,\,\,\,\,\,\,\,return} $\sum_{s'=1}^{N}
            x(s')\left[\ln{x(s')} - \ln{y}\right]$
        \STATE {\bfseries }\textbf{\,\,\,\,\,\,end function}

        \FOR{$t=0,\dotsc,\infty$}
            \STATE Receive $s_t$, choose action $a_t$.
            \STATE $L_0 := L_0 - f_0(m', n', s_t)$.
            \STATE $L_1 := L_1 - f_1(m, n, s_t, a_t)$.
            \STATE Update $m, m', n, n'$.
            \STATE $L_0 := L_0 + f_0(m', n', s_t)$.
            \STATE $L_1 := L_1 + f_1(m, n, s_t, a_t)$.
            \STATE Compute $L =-2(L_0 - L_1)$.
        \ENDFOR
    \end{algorithmic}
\end{algorithm} 

Source code for the RLAPSE framework including implementations of (i) sparse matrices to store the 
transition counts and (ii) incremental computation of the test statistic $L$ is publicly 
available~\cite{rlapse_repo}.
}

\subsection{Regret bound for RLAPSE}
Recall that regret is an RL algorithm performance metric as defined in \eqref{eq: regret}. Performance bounds on algorithms $\A_i$ are typically available for an environment with model $M_i$ they were designed for, whereas equivalent bounds for $\A_i$ applied to other environment structures will be hard to establish; e.g., regret bounds for LinUCB applied to a CMAB environment are given in~\cite{li2010contextual}, but of course no analysis for the case of a controlled MDP is performed in this work. Hence, the main ingredient of showing that the regret of RLAPSE is asymptotically not worse than using the more complex MDP-based RL algorithm exclusively is Theorem~\ref{thm:type2}. This theorem establishes that the probability of using $\A_0$ (yielding a myopic policy) if the environment is not an uncontrolled MDP (model $M_0$) exponentially decays to 0 as the number of interactions grows.  

In order to quantify the efficacy of the LR test for uncontrolled versus controlled MDP, we derive an upper bound for the probability of a Type 2 error\footnote{A Type 2 error is defined as the non-rejection of a false null hypothesis.}, showing that this probability goes to zero at an exponential rate whenever the environment is a controlled MDP and exploration occurs. In other words, the test will (eventually) correctly identify any controlled MDP environment.  

For a given MDP $M$, define $
\theta = \max_{i,j} \left(\overline{P}_{ij} - \underline{P}_{ij}\right)$; %
$\theta$ and $\rho(M,Y)$ are related through $\theta/2 \le \rho(M,Y) \le N \theta$ for any $Y\in\MCs(M)$. The null hypothesis of the LR test for uncontrolled versus controlled MDP can then be restated as $H_0 : \theta = 0$, and the alternate hypothesis
is $H_1 : \theta > 0$. A Type 2 error occurs if $H_0$ is accepted when $H_1$ is correct. The probability of a Type 2 error at significance level $\alpha$ after $T$ steps is 
    \begin{align}\label{def:beta}
        \beta(T) = &\P\left(L \le t \, \text{up to time $T$} \,|\, H_1\right), \\
        \text{ where} & \,t = \chi^2_{1 - \alpha, df},  \text{ and } df = N (A-1) (N-1). \notag
\end{align}
For a homogeneous combined system of MDP and policy with nonzero exploration rate, we will show that $\beta(T)$ converges to zero exponentially as $T \rightarrow \infty$. The policy $\pi^{(E)}$ is specified by
\be\label{def:q}
\pi^{(E)}_a(s) = \frac{\varepsilon}{A} + (1-\varepsilon) \, \pi_E(a\given s),
\ee
where $\varepsilon$ is the exploration probability and $\pi_E$ is the exploitation policy which selects actions to maximize the current estimate of the action-value function.
The decay rate of $\beta$ is defined as
\be\label{def:kappa}
\kappa^* = \sup \{\kappa \,:\, \lim_{T \rightarrow \infty} e^{\kappa T} \, \beta(T) = 0 \}.
\ee

\begin{theorem}\label{thm:type2}
    $\beta$ decays exponentially to zero as $T \rightarrow \infty$ for all $\theta > 0$ and all $r > 0$. The decay rate satisfies the lower bound
    \be
    \kappa^* \ge c \, \varepsilon^2 \, \theta^2 \, P_\text{min}^2 \, \underline{w_{I}}^2,
    \ee
    where $P_{\text{min}}$ is the smallest nonzero entry of $P(j | i, a)$, $\underline{w_{I}}$ is the smallest component of the Perron vector $w_I$ of the induced MC $P_E(j | i) = \sum_{a} \pi^{(E)}_a(i) \, P(j | i,a)$, and
    \bee
    c = \left(2 A N (24 A)^2\right)^{-1} \, \min \bigg\{1 , \frac{A N \, (1 - \tau_1(P_E))^2}{4} \bigg\}.
    \eee
    
    See Section~\ref{ssec:sup:type2} of the appendix for details and proof.
\end{theorem}

\begin{corollary}\label{cor:regret}
    Let $\A_0$ and $\A_1$ denote RL algorithms, and assume regret bounds $R^0_{\text{u}}$, $R^1_\text{u}$, and $R^1_\text{c}$ are known, where $R^i_\text{u}$ and $R^i_\text{c}$ denote the regret of $\A_i$ applied in an uncontrolled and controlled MDP environment, respectively. Then, the expected regret of using RLAPSE with $\A_0$ and $\A_1$, and confidence level $\alpha$ as $T\rightarrow\infty$ satisfies %
    \begin{equation*}
        \E \{R(T)\} = 
        \mathrm{O}\Bigl( \alpha R^1_\text{u}(T) + (1-\alpha)R^0_\text{u}(T) \Bigr)
    \end{equation*}
    if the environment is an uncontrolled MDP, and 
    \begin{equation*}
        \E \{R(T)\} =          
        \mathrm{O}\left(R^1_\text{c}(T)\right)
    \end{equation*}  
    if the environment is a controlled MDP.
\end{corollary}
\begin{pf}
    The case for an uncontrolled MDP is clear because the probability of rejecting the true null hypothesis and using $\A_1$ is $\alpha$. For a controlled MDP, we note that even though there is no regret bound $R^0_\text{c}$ assumed (and will typically not be available), we definitely have $R^0_\text{c}(T)\leq \bar{r} T$ for any interval $[0,T]$ during which $\A_0$ is in operation, whereas, according to~Theorem~\ref{thm:type2}, the probability of this event (a Type 2 error) decays to 0 \emph{exponentially} in $T$; the details are explained in Section~\ref{ssec:sup:regret} of the Supplementary Material.
\end{pf}

\section{Evaluation}\label{sec: eval}
In this section, we evaluate the performance of the RLAPSE framework: first on a constructed set of 
example environments; then on randomly generated MDP environments. The constructed example scenarios are 
based on the broker-supplier setting described in Section \ref{sec: intro}, demonstrating the potential for this problem 
to manifest as either a controlled or uncontrolled MDP as the result of subtle changes to the 
environment. The experiments for randomly generated environments demonstrate that our results apply more 
generally. For the purpose of evaluation, we select two specific RL algorithms to test RLAPSE on environments with different structures. Specifically, we implement two variants of Q-learning as 
analyzed in~\cite{even2003learning}, with $\omega=0.7$ and a constant exploration probability of $0.2$. 
We set $\gamma=0$ for the myopic algorithm $\A_0$ for uncontrolled MDPs, and $\gamma=0.9$ for $\A_1$, 
the algorithm for controlled MDPs. In order to collect enough statistics for the LR test using MDPs of 
different size, the orchestrator has to be started at different $T_0$. Thus, $T_0 = N^2A$ 
was chosen as the minimum amount of data before performing the first test.

The source code for the experiments presented is available on GitHub~\cite{rlapse_repo}.

\subsection{Toy broker example}
We now return to the example presented in Figure~\ref{fig: broker} of a broker selecting from a set 
of suppliers. While this example is simple in nature, it serves to illustrate clearly the potential 
for ambiguity when a designer is faced with a new MDP environment and must select an appropriate 
algorithm for an RL agent.  

We model the broker decision problem as an MDP. The broker periodically purchases units of a commodity 
from a supplier, chosen from a set of $D$ suppliers. In other words, the agent selects an action
$a_t = i \in [D]$ at each time step $t$. The length of the time interval between these purchases depends on the 
context, e.g., this could be a matter of seconds in a high-frequency trading scenario, or days in the case of purchasing batches of a new drug or vaccine as it is manufactured. 
Nonetheless, the broker must choose a supplier at each such time step. We assume that each supplier $i$ 
can offer a unit of the commodity at one of $K$ prices $b_i \in \{1, 2, \dots, K\}$. The state of the 
MDP can then be represented equivalently as $s = (b_1, b_2, \dots, b_{D})$ or 
$s=b_1 + b_2K + \dotsb + b_{D}K^{D-1}$. The reward function is monotonically decreasing in the 
price offered by the chosen supplier:
\begin{equation}\label{eq:reward_mono}
    b_i < b'_i \quad\Longrightarrow \quad
        r( s, i) \geq r( s', i), 
\end{equation}
where $b'_i$ is the $i^{th}$ element of the state $s'$. Other factors which could lead the broker to 
favor one supplier over another, and which can be reflected in the reward function accordingly, include 
the quality of the commodity offered by each supplier or of the service provided in delivering it.  

To simplify the modeling process, assume that suppliers change their offered price only in steps of 1: 
if a supplier at time $t$ offers price $b$, then at time $t+1$ it offers one of $\{b-1,b,b+1\}$. Assume 
also that a supplier's state transitions are dependent only on whether or not it is used, not on which 
alternative supplier is used and what the states of the other suppliers are. Associate with each 
supplier $i$ four parameters $p_{+,i}$, $p_{-,i}$, $q_{+,i}$, $q_{-,i}$, where
\begin{itemize}
    \item $p_{+,i}$ is the probability of supplier $i$ increasing its price by 1 if it is used, that is
    $
        p_{+,i} = P( b'_i=k+1 \given b_i=k, a=i)
    $;
    \item $q_{+,i}$ is the probability of supplier $i$ increasing its price by 1 if it is \emph{not} 
    used, i.e.,
    $
        q_{+,i} = P( b'_i=k+1 \given b_i=k, a\neq i)
    $;
    \item analogously, $p_{-,i}$ corresponds to supplier $i$ decreasing its price when 
        it is used, and $q_{-,i}$ corresponds to supplier $i$ decreasing its price when 
        it is \textit{not} used; 
    \item $(1-p_{+,i}-p_{-,i})$ corresponds to supplier $i$ maintaining its price when 
        it is used, and $(1-q_{+,i}-q_{-,i})$ corresponds to supplier $i$ maintaining its price when 
        it is \textit{not} used; finally 
    \item if $b_i=1$, then we have to set $p_{-,i}=q_{-,i}=0$, and if $b_i=K$, then we have to set 
        $p_{+,i}=q_{+,i}=0$.
\end{itemize}
Now we can parametrize the transition probability tensor $P(:)$: $P(s' \given s,a) = 0 $ 
if $|s'_j-s_j|>1$ for any $j$, and else
\begin{align*}
    P(s' {\given}& s,a) = \\
    &\psi_{a} \prod_{j\in U_-} q_{-,j} \prod_{j\in U_+} q_{+,j} 
       \prod_{j\in U} (1-q_{+,j}-q_{-,j}),
\end{align*}
where $U_-,U_+,U$ denote the sets of indices $j\neq a$ so that the corresponding suppliers decrease, 
increase, do not change their prices, respectively; and $\psi_{a}$ equals $p_{+,a}, p_{-,a}$ or 
$(1-p_{+,a}-p_{-,a})$, depending on whether the chosen supplier $a$ increases, decreases or maintains 
its price, i.e., whether $b'_a=b_a+1$, $b'_a=b_a-1$ or $b'_a=b_a$. 

The rewards depend on the performance and price of the chosen supplier, and not on the other (unused) suppliers, 
and are subject to the monotonicity constraint~\eqref{eq:reward_mono}. Thus, for every supplier $i$, we 
introduce $K$ parameters $r_{1,i}\geq r_{2,i}\geq \dotsb \geq r_{K,i}$, where $r_{b_a, a} = r(s,a)$.  

We begin our evaluation with a set of two concrete examples fitting the broker model, one 
for the uncontrolled MDP case, and one for the controlled MDP case. It is important to note that there is no 
way of knowing from the outset which of these models most accurately captures the true environment. For 
each of the two examples, we show how a broker performs when it assumes the environment is uncontrolled 
(controlled) and employs algorithm $\A_0$ ($\A_1$) and when it uses our RLAPSE framework to 
make the choice automatically. We initially use a small state-action space with two suppliers and two 
price categories to allow easily explainable results, and subsequently provide more general results for larger 
state-action spaces in the following subsections. 

\subsubsection{Uncontrolled MDP}
The uncontrolled MDP case represents a typical baseline assumption about an environment like the one 
in our broker example. In this case, the prices offered by suppliers are determined by economic factors (e.g., the overall market demand
for the commodity in question), which determine the transition probabilities of the MDP. Thus, the purchasing decisions of an individual broker have no effect on 
state transition probabilities. 

\begin{figure}[tbh]
    \centering
    \includegraphics[width=\mainfigwidth]{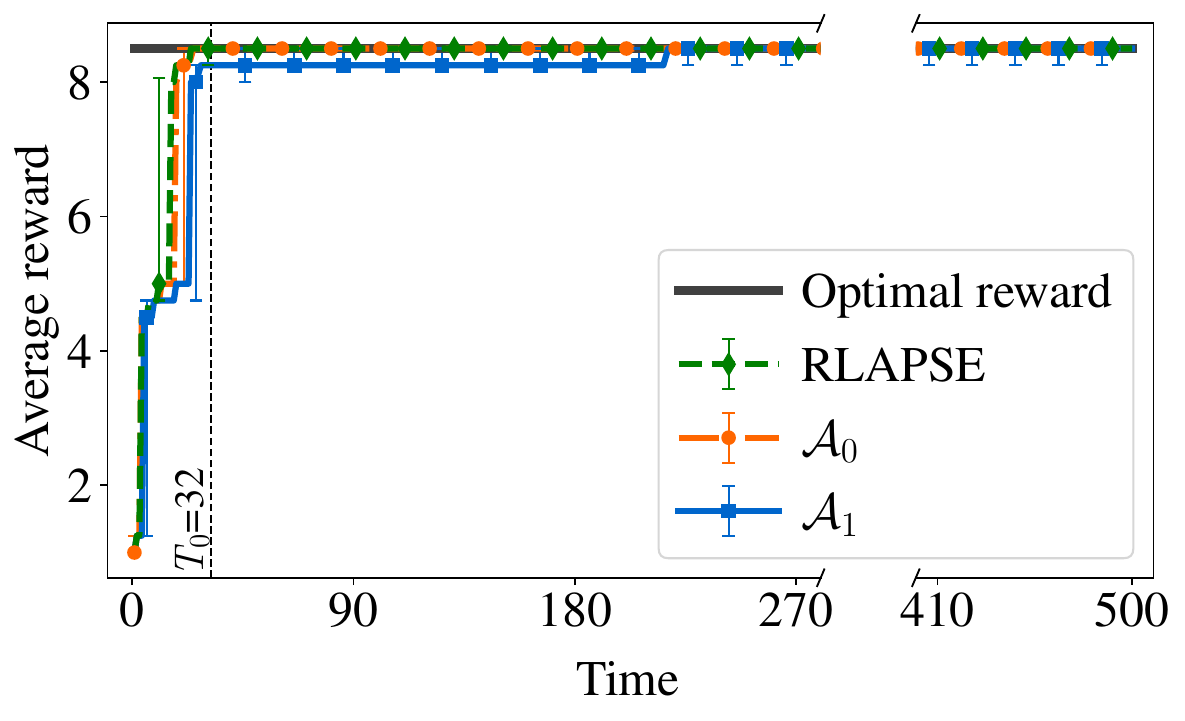}
    \includegraphics[width=\mainfigwidth]{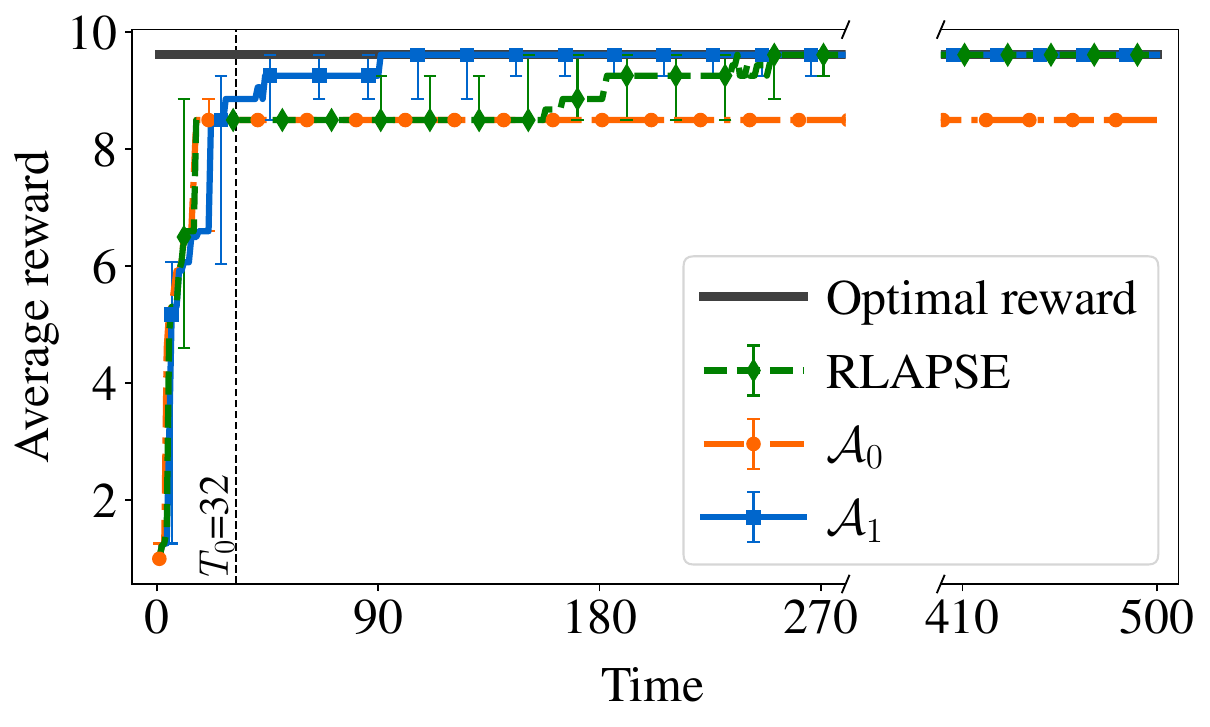}
    \caption[na]{Comparison of $\A_0$, $\A_1$ and their combination in an RLAPSE framework for the
        broker example with 2 suppliers, each with 2 price categories, in the uncontrolled MDP 
        (\textbf{Top}) and the controlled MDP environment (\textbf{Bottom}). Note the split time axes. 
        The error bars correspond to $1^{\text{st}}$ and $3^{\text{rd}}$ quartiles for the median value 
        of average reward obtained from 100 different realizations of the experiment.
    }
    \label{fig:toy_broker}
\end{figure}

Consider a simple case with 2 suppliers, each with 2 price categories. Since the environment is an 
uncontrolled MDP, in this case we choose $p_{+,i} = q_{+,i} = p_{-,i} = q_{-,i} = 0.5$ for both suppliers. 
Suppose supplier $1$ offers a far superior commodity all of the time, hence the immediate reward is always 
better for choosing this supplier. Supplier $2$, meanwhile, offers a poor quality commodity. Take the 
rewards to be:
\begin{align*}
r_{1,1} &= 15, & r_{2,1} &= 2, & r_{1,2} &= r_{2,2} = 1.
\end{align*}
It is clear that the best immediate reward can be gained by the myopic policy of always choosing supplier
$1$ (the first column of the rewards matrix). As this MDP is uncontrolled, the myopic policy turns out
to be optimal in terms of gain as well. The performance of $\A_0$, $\A_1$ and RLAPSE in this environment 
is depicted in Figure~\ref{fig:toy_broker}(Top), from which it can be observed that RLAPSE selects 
$\A_0$ almost exclusively, hence their performance is identical, whereas $\A_1$ also converges to 
the optimal policy, but slower.

\subsubsection{Controlled MDP}
On the other hand, latent behavioral characteristics of the suppliers in this environment can easily give rise to a 
controlled MDP. For example, if a supplier gathers data on each of its customers and runs an algorithm 
seeking to maximize its own returns by adapting prices to the measured demands, the environment becomes 
a controlled MDP. The supplier could either tend to increase the price offered 
to the broker when a purchase is made, in an attempt to take advantage of the broker, or could tend to reduce its price when the broker purchases from elsewhere, in a bid to 
attract the broker's patronage. 

Suppose supplier $1$ in the above example adapted its prices in one such way, so that the probability
of dropping its price when the broker purchases elsewhere is increased. Consider what happens when we fix all 
parameters as before, and simply decrease the parameter $q_{-,1}$ from $0.5$ to $0.3$, indicating that 
supplier $1$ is more inclined to drop its price when the broker purchases from other suppliers. It turns out that 
as long as $\frac{p_{-,1}}{q_{-,1}}>\frac{r_{1,1}+r_{2,1}-2r_{1,2}}{r_{1,1}-r_{2,1}}$, the myopic 
policy becomes suboptimal in this case. We can see the effect of this small parameter change in this 
example by once again comparing the performance of a myopic policy against a more complex algorithm. 
Figure \ref{fig:toy_broker}(Bottom) illustrates the results of this experiment. 
As it can be seen, RLAPSE successfully selects $\A_1$ after time $T_0$, and also converges to the optimal 
policy as well as $\A_1$ alone. On the contrary, $\A_0$ converges only to a sub-optimal policy in this 
environment. 

\subsection*{Simulation results}
\begin{figure}[tbh]
    \centering
    \includegraphics[width=\mainfigwidth]{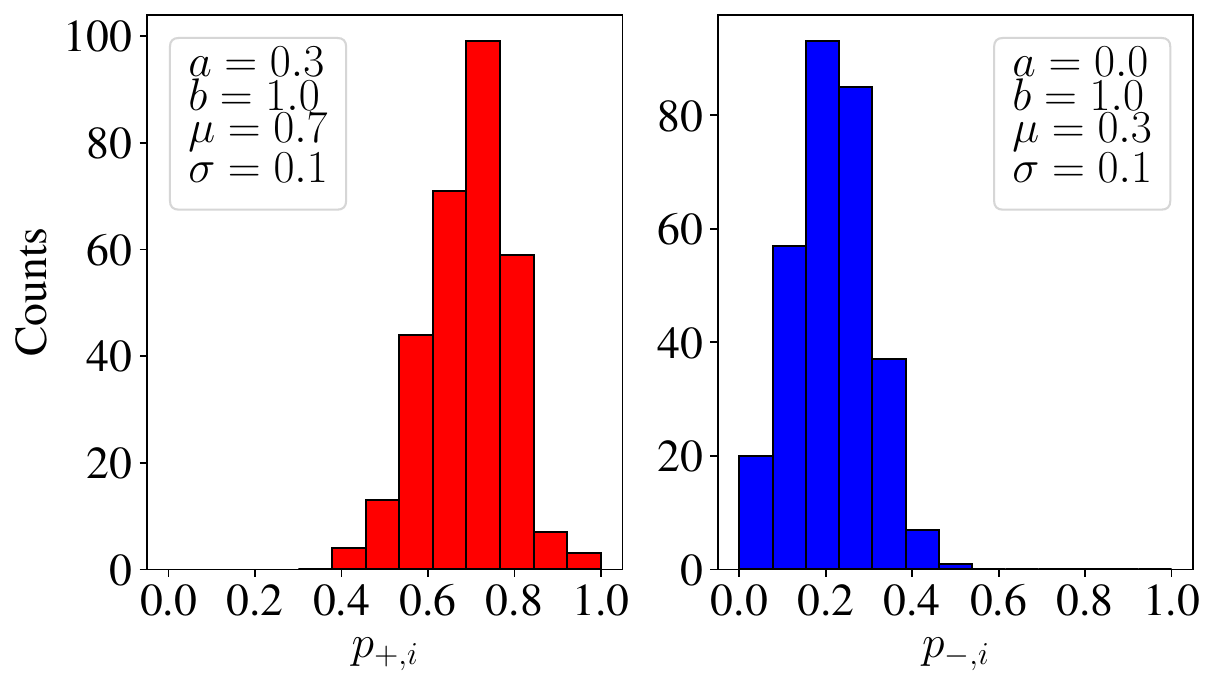}
    \caption{
        Histograms of samples for $p_{\star,i}$ drawn from the truncated normal distribution with 
        parameters given in the legends for $\epsilon=0.3$. The parameters $q_{\star,i}$ only differ 
        by $\epsilon$.
    }
    \label{fig:p_hist}
\end{figure}

\begin{figure}[tbh]
    \centering
    \includegraphics[width=\mainfigwidth]{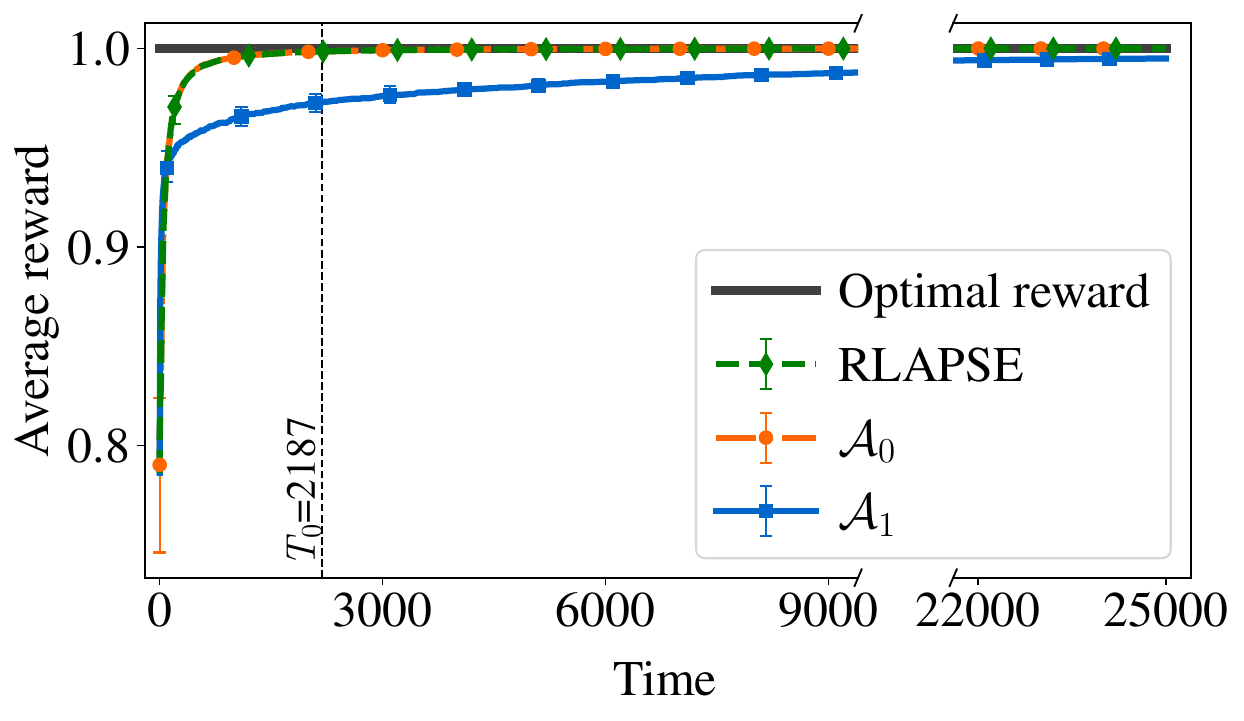}
    \includegraphics[width=\mainfigwidth]{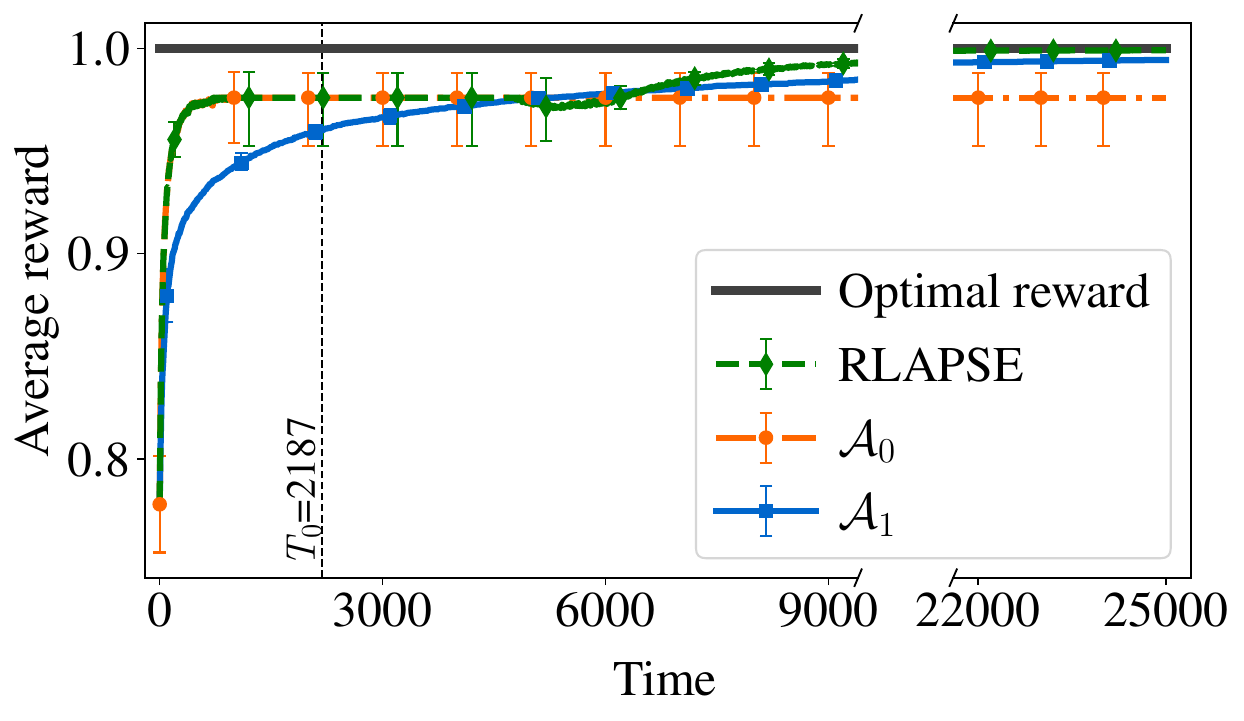}
    \caption{
        Comparison of RLAPSE, $\A_0$ and $\A_1$ algorithms for $D = 3$, $K = 3$,
        $\epsilon = 0$ (\textbf{Top}) and $\epsilon = 0.4$ (corresponding to a controlled MDP environment, 
        \textbf{Bottom}) in the broker example.
        The error bars correspond to $1^{\text{st}}$ and $3^{\text{rd}}$ quartiles for the median. 
    }
    \label{fig:larger_broker}
\end{figure}

First, we evaluate the performance of RLAPSE using the described broker example with randomly 
generated data. Specifically, we randomly generate 100 MDPs (with 100 different realizations for each) 
corresponding to $D=3$ suppliers, each with $K=3$ price categories (thus, $N = 27$), with various values of 
$\epsilon = p_{+,i}-q_{+,i} = q_{-,i}-p_{-,i}$ for all $i$; $p_{+,i}$ and $p_{-,i}$ were drawn from 
$\tN(0.7,0.1;0.3,1.0)$\footnote{$\tN(\mu,\sigma;a,b)$ denotes the truncated normal distribution on the interval 
$[a,b]$ with mean $\mu$ and variance $\sigma^2$.} and $\tN(0.3,0.1;0.0,1.0)$, respectively
(the corresponding histograms for $\epsilon = 0.3$ are shown in Figure~\ref{fig:p_hist}). The results of 
the simulations using these settings are depicted in Figure~\ref{fig:larger_broker}.
As can be observed in Figure~\ref{fig:larger_broker}(Top), RLAPSE almost exclusively selects $\A_0$,
and thus performs as well as $\A_0$ alone in this uncontrolled MDP environment. In the controlled MDP
environment, see Figure~\ref{fig:larger_broker}(Bottom), $\A_0$ is not able to converge to the optimal policy even though it has 
a higher initial convergence rate than that of $\A_1$. RLAPSE, however, successfully selects $\A_1$ 
(once the LR test starts rejecting $H_0$ at a high rate) and converges to the optimal policy.  

\begin{figure}[tbh]
    \centering
    \includegraphics[width=\mainfigwidth]{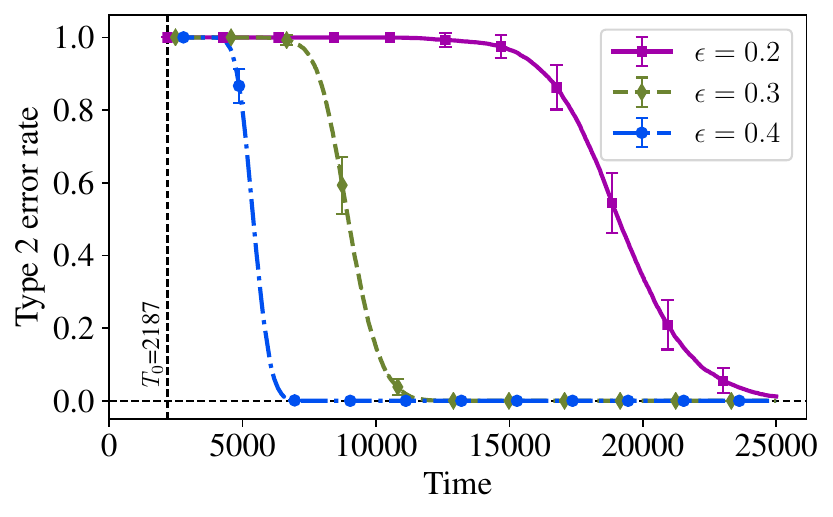}
    \caption{Type 2 error rate for various values of $\epsilon$ in the broker example. 
        The error bars represent the 99\% confidence interval for the mean.
    }
    \label{fig:broker_t2err}
\end{figure}

Next, we test the main assertion of Theorem~\ref{thm:type2}, namely that the probability of Type 2 
errors decays exponentially, and we show the intuition that the decay rate depends on how ``different'' 
from an uncontrolled MDP the environment is.
The smaller $\epsilon$, the smaller the effect of the action $a_t$ on the state transition, and hence 
we expect the frequency of Type 2 errors to decrease at an exponential rate, with the rate increasing 
as $\epsilon$ increases. Figure~\ref{fig:broker_t2err} illustrates this for $\epsilon\in\{0.2,0.3,0.4\}$.

\subsection{Randomly generated MDPs}
For the next set of experiments, we simply generated random MDPs 
for $N=\{10,50\}$ states and $A=3$ actions. These experiments demonstrate that our assertions hold 
outside of the constraints placed on the environment in the broker example.
Transition probabilities were drawn from a Gamma distribution (with shape and scale being $1$ and $5$, 
respectively) and then normalized; the entries of the reward matrix were also drawn from a Gamma 
distribution (shape and scale equal $0.1$ and $5$, respectively). A total of 100 MDPs (with 100 different 
realizations for each MDP) for every of these four environment structures were generated:
\begin{enumerate}[(I),nosep,align=left]
    \item $p(s'|s,a) = p(s')$, i.e., the states are i.i.d.;
    \item $p(s'|s,a) = p(s'|s)$, i.e., the MDP is uncontrolled, with all pages being equal to each other, but of rank $\geq$ 1;
    \item $p(s'|s,a) = p(s'|a)$, i.e., the MDP is controlled, but all transition matrices are rank-1; 
        and 
    \item the general case of $p(s'|s,a)$ with no specific structure.
\end{enumerate}
The parameter $T_0$ of the orchestrator and the parameters of the Q-learning algorithms were set to the 
same values as it is described in the broker example above.

\begin{figure}[tbh]%
    \centering
    \includegraphics[width=\mainfigwidth]{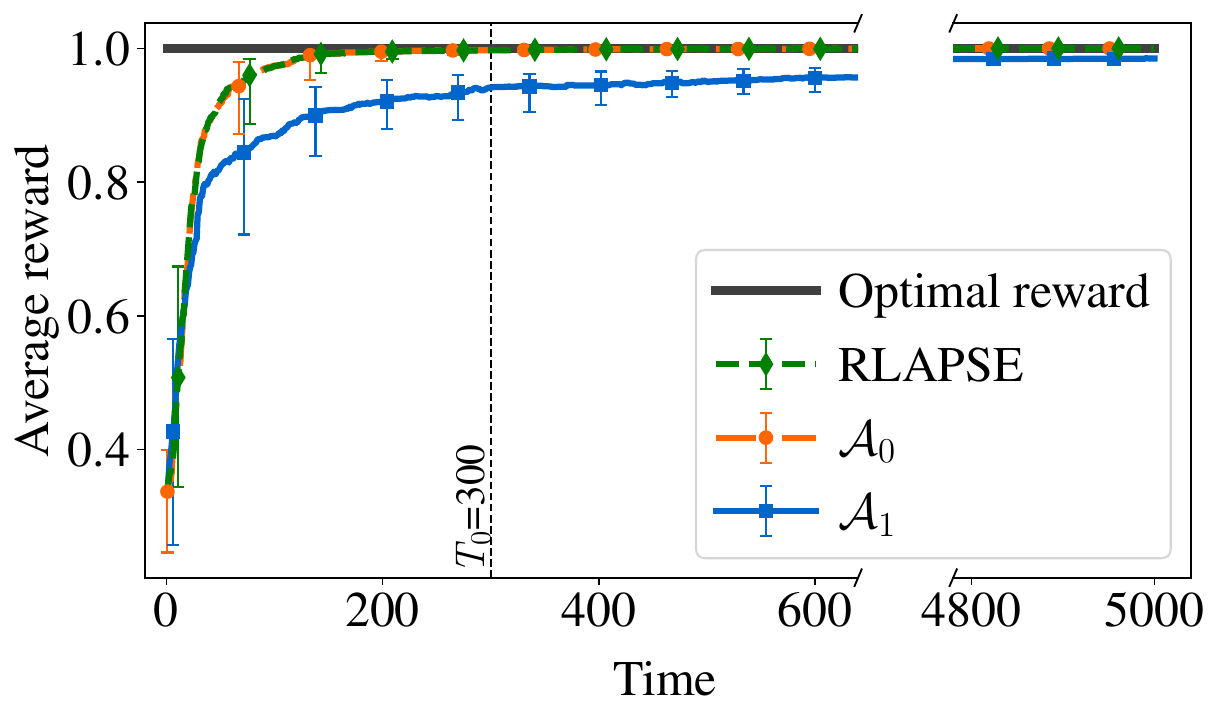}
    \includegraphics[width=\mainfigwidth]{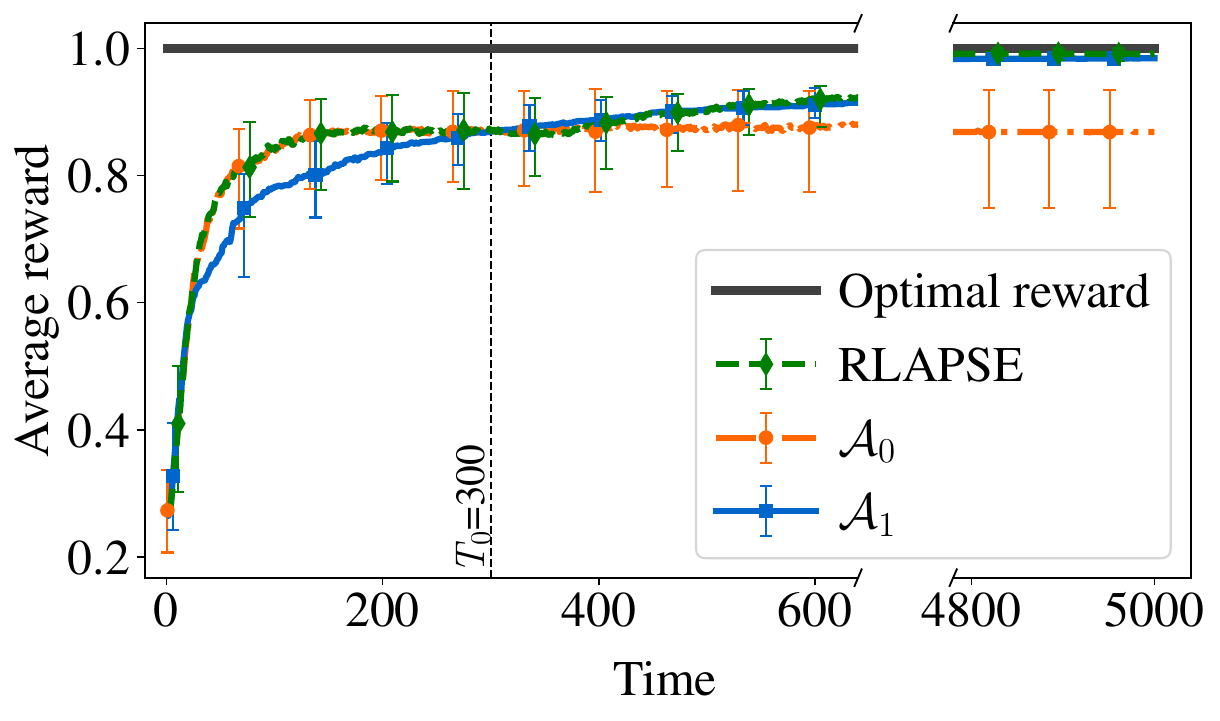}
    \caption[na]{Comparison of $\A_0$, $\A_1$ and their combination in the RLAPSE framework in 
        environments of structure (I)---\textbf{Top} and (III)---\textbf{Bottom} for MDPs with $N=10$
        states and $A=3$ actions.
        The error bars correspond to $1^{\text{st}}$ and $3^{\text{rd}}$ quartiles for the median. 
    }
    \label{fig:env13_s10}
\end{figure}

\begin{figure}[tbh]%
    \centering
    \includegraphics[width=\mainfigwidth]{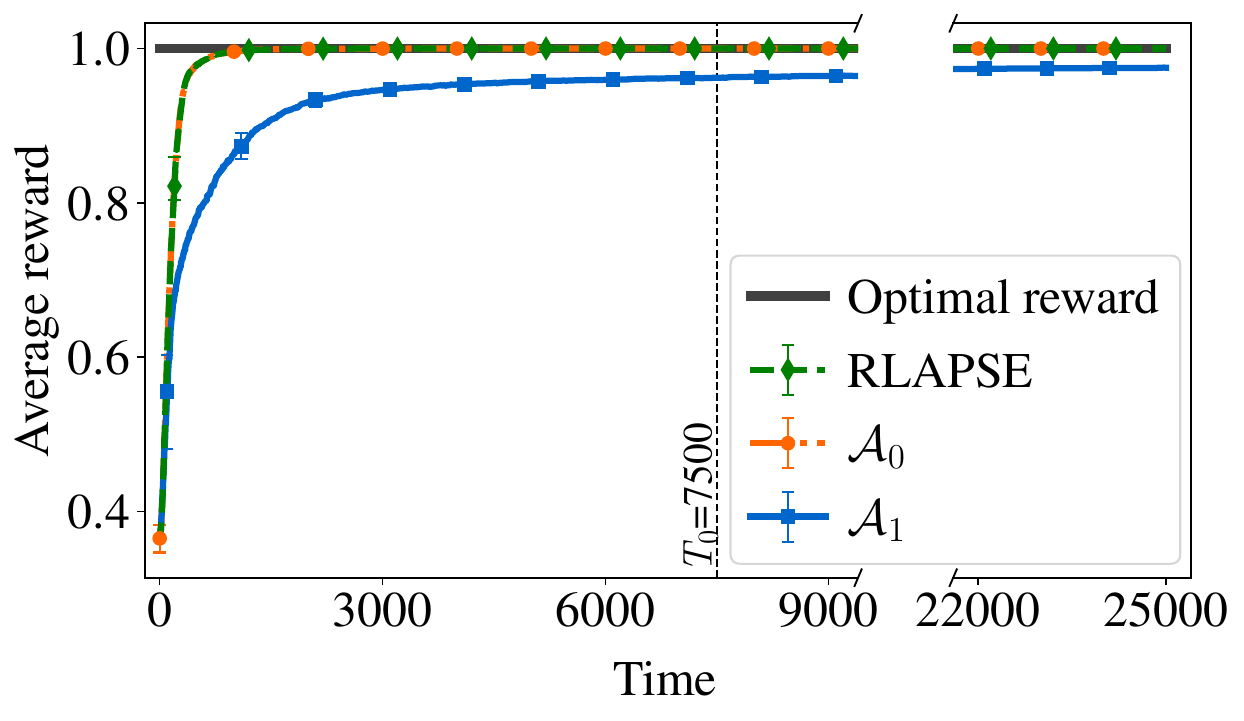}
    \includegraphics[width=\mainfigwidth]{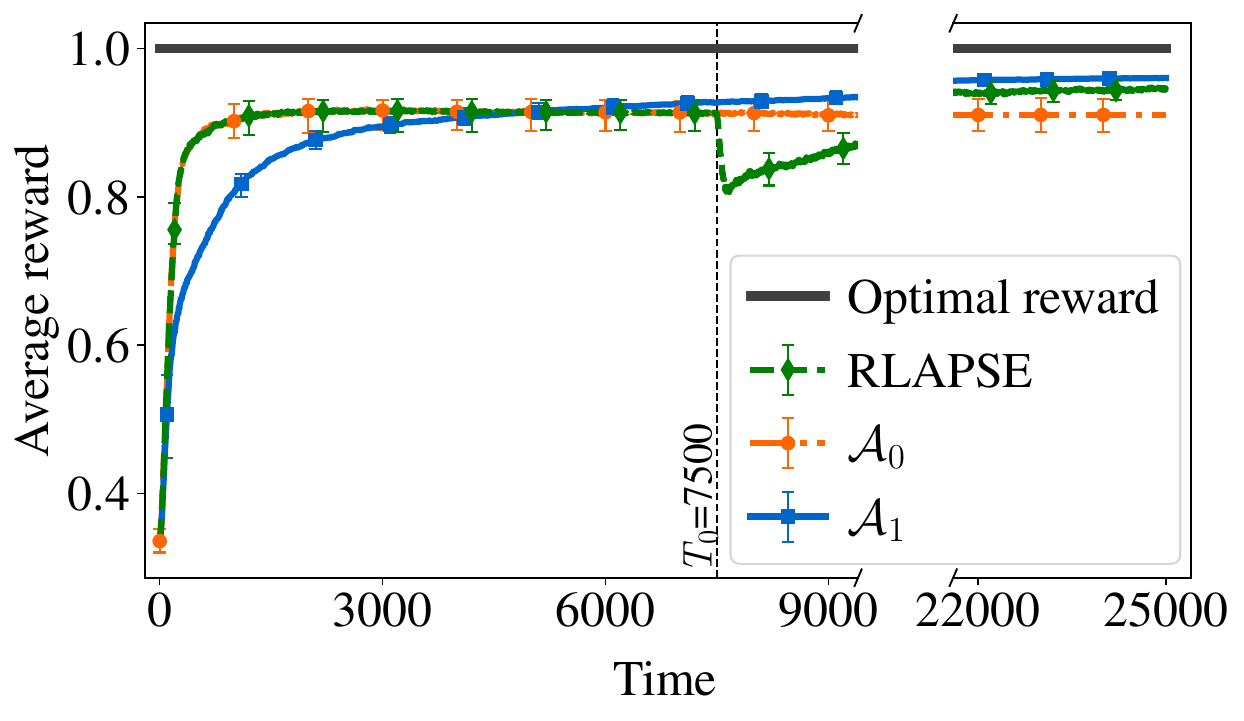}
    \caption[na]{Comparison of $\A_0$, $\A_1$ and RLAPSE in uncontrolled MDP (\textbf{Top}) 
        environments of structure (II) and controlled MDP (\textbf{Bottom}) environments of structure 
        (IV) for MDPs with $N=50$ states and $A=3$ actions.
        The error bars correspond to $1^{\text{st}}$ and $3^{\text{rd}}$ quartiles for the median. 
    }
    \label{fig:env24_s50}
\end{figure}

The results of the comparative performance of the algorithms $\A_0$, $\A_1$ and their combination in 
RLAPSE are depicted in Figure~\ref{fig:env13_s10}---for environments of structure (I) and (III) for 
MDPs with $N=10$ states, and in Figure~\ref{fig:env24_s50}---for environments of structure (II) and (IV) 
for MDPs with $N=50$ states. As can be observed, RLAPSE selects $\A_0$ almost exclusively for 
environments of structure (I) and (II), hence their performance is nearly identical,
whereas $\A_1$ also converges to the optimal policy, but slower, in these uncontrolled MDP 
environments (see top panels). In the controlled MDP environments of structure (III) and (IV), 
algorithm $\A_0$ (due to its myopic nature) is not able to converge to the optimal policy. RLAPSE, 
however, successfully selects $\A_1$ after time $T_0$ and converges to the optimal policy as well as 
$\A_1$ alone (see bottom panels).

Table~\ref{tab:s50} illustrates the statistics of the LR test for MDPs with $N = \{10, 50\}$ states 
(top and bottom, respectively) and $A = 3$ actions. For each of the four environment structures, we 
collect how often the null hypothesis (corresponding to an environment of structure (I) or (II)) 
was accepted or rejected. The observations in this table, along with those in Figures~\ref{fig:env13_s10} 
and~\ref{fig:env24_s50}, validate our expectations about RLAPSE with the LR test: in most cases, 
the RLAPSE framework accepted the null hypothesis for structures (I) and (II), and rejected the 
hypothesis otherwise, i.e, for structures (III) and (IV).
\begin{table}[bth]
    \centering
    \begin{tabular}{r||c|c|}
        Env. struc. & Hyp.\ accepted   & Hyp.\ rejected \\\hline
        (I)     & 4679.23 (99.56 \%) & 20.77 (0.44 \%)   \\
        (II)    & 4667.12 (99.30 \%) & 32.88 (0.70 \%)   \\
        (III)   & 74.14 (1.58 \%)    & 4625.86 (98.42 \%)\\
        (IV)    & 32.04 (0.68 \%)    & 4667.96 (99.32 \%) \\
        \multicolumn{1}{c}{} & \multicolumn{1}{c}{}\\
        Env. struct. & Hyp.\ accepted & Hyp.\ rejected \\\hline
        (I)     & 17500 (100 \%)   & 0 (0 \%) \\
        (II)    & 17500 (100 \%)   & 0 (0 \%)\\
        (III)   & 582.21 (3.33 \%) & 16917.79 (96.67 \%)\\
        (IV)    & 162.88 (0.93 \%) & 17337.12 (99.07 \%) 
    \end{tabular}
    \caption{
        Results of the LR test, collected over 100 examples of random MDPs with $N=10$ states, 
        $T-T_0 = 4700$ (\textbf{Top}) time steps and for MDPs with $N=50$, $T-T_0 = 17500$ 
        (\textbf{Bottom}) for each of the 4 environment structures.
    }
    \label{tab:s50}
\end{table}

\begin{figure}[tbh]%
    \centering
    \includegraphics[width=\mainfigwidth]{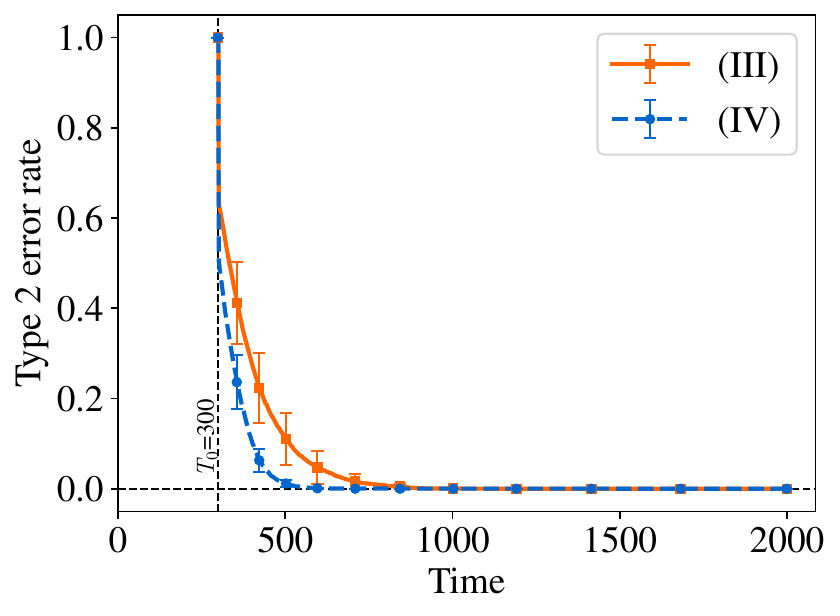}
    \includegraphics[width=\mainfigwidth]{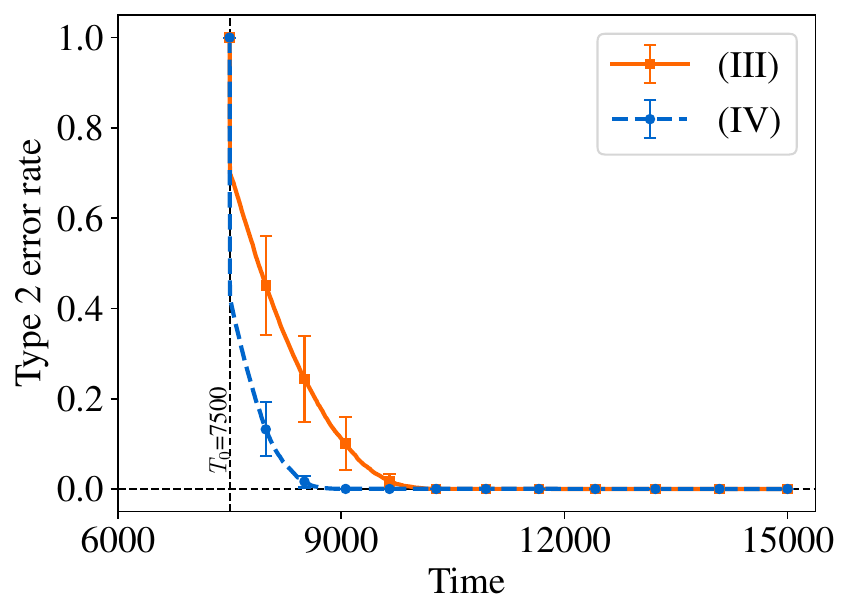}
    \caption[na]{Type 2 error rate for MDPs with $N = 10$ (\textbf{Top}) and $N = 50$ (\textbf{Bottom}) states.
        The legends indicate what structure of environment the data is obtained for.
        The error bars represent the 99\% confidence interval for the mean.
    }
    \label{fig:t2err_s10_s50}
\end{figure}

Figure~\ref{fig:t2err_s10_s50} depicts the decay over time of the rates of Type 2 errors for 100 
randomly generated MDPs and 100 different realizations per each of them. Recall that the Type 2 error 
decays to zero as time goes to infinity (the decay, once it starts, is at least exponential), 
hence most Type 2 errors are made early and stop occurring as more data is collected.
 
\section{Conclusions}\label{sec: conc}
In summary, we have confronted an important issue facing designers of RL systems, namely how to deal with uncertainty in the environment structure, and hence in the choice of learning algorithm. We have distinguished between environments in which an RL agent's actions influence future states, and environments in which they do not. We have shown that an environment fitting the latter description is a sufficient condition for a myopic RL algorithm to find an optimal policy. Based on this fact, we presented the LR orchestrator, a mechanism which continuously tests the environment for this condition and selects an appropriate algorithm accordingly. The orchestrator, combined with a selection of two RL algorithms, constitutes the RLAPSE framework which switches to the appropriate algorithm based on the environment structure detected. We provided a regret bound for this RLAPSE framework.  

In terms of validation, we presented an example of an environment based on a broker in which RLAPSE offers clear advantages. We then presented experiments which demonstrate that RLAPSE is effective in more general unknown MDP environments.  
 
\bibliographystyle{plain}

\cleardoublepage
\onecolumn
\appendix
\section{Appendix}
\subsection{Worst-case bounds on performance gap}\label{ssec:sup:perfgap}
We will need one additional result using $\tau_1$, namely a theorem bounding the sensitivity of the Perron vector to perturbations in a row-stochastic matrix:
\begin{theorem}[%
    \protect{\cite[Thm.~2]{Seneta1991}}\footnote{Note that \cite{Seneta1991} contains a typo, namely 1-norms are written instead of $\infty$-norms.}%
    ]\label{thm:seneta} 
    Let $P$ be the transition matrix of a unichain and scrambling MC, and let $E$ be such that $P+E$ is also the transition matrix of a unichain MC. Let $w$ and $w'$ denote their Perron vectors. Then,
    \begin{equation}
        \| w - w' \|_1 \leq \frac{1}{1- \tau_1(P)} \|E\|_\infty.
    \end{equation}
\end{theorem}
The maximum gain $g_\pi$ is achieved by at least one Markov randomized policy, and hence we may restrict our attention to such policies.  The gain has a simple expression in terms of the induced Markov chain; for the proof of this lemma, see Chapter 8 of~\cite{puterman1994markov}.
\begin{lemma} \label{lem:gain}
    Assume that $\pi$ is an MR policy so that the induced $P_{\pi}$ is unichain. Then, the gain satisfies $\gamma_\pi = g_\pi \one$ for some scalar $g_\pi$. With $w_\pi$ the Perron vector of $P_{\pi}$, we  have
    \begin{equation}\label{eq:gpi}
        g_\pi = w_\pi\trans r_\pi.
    \end{equation}
\end{lemma}

Assume that $M$ is \emph{unichain}, and we know that for an MDP $M=(P(:\nolinebreak), \varpi, R)$, the set $\MRPs(M)$ contains at least one MRP corresponding to a policy $\pi^*$ with the optimal average reward. 
Then, given $\pi^*$, the optimal average reward is given by 
\[
g^* \coloneq g_{\pi^*} = w^* r_{\pi^*},
\]
where $w^*\coloneq w_{\pi^*}\trans$ is the Perron vector of $P_{\pi^*}$.

The performance gap between the optimal policy $\pi^*$ and the myopic policy $\pi^C$  is bounded from above by the difference between the gains of the myopic policy and \emph{any} other MR policy (in fact, this bound equals the performance gap, otherwise the optimal policy would not be optimal). We will now compute one such bound. 
For a given policy $\pi'$, denote the corresponding transition matrix by $Y = P_{\pi'}$. Then we can write any other $P_\pi \in \MCs(M)$ as $P_\pi = Y+E$, where
\[
E_{ij} = \sum_{a\in A} P(j \given i, a)  (\pi_a(i) - \pi'_a(i)).
\]
We now bound $\|E\|_\infty$. For convenience, define the matrices $\overline{P}_{ij} = \max_a P(j \given i, a)$ and $\underline{P}_{ij} = \min_a P(j \given i, a)$, $\widetilde{P}_{ij} = (\overline{P}_{ij} + \underline{P}_{ij})/2$. Then, we have 
\begin{multline}\label{eq:Einfty:sup}
    \|E\|_\infty = \max_i \sum\limits_j \left| \sum\limits_{a\in A} P(j \given i, a)  ({\pi}_a(i) - \pi'_a(i)) \right| \\
    \leq 
    \max_i \sum\limits_j \frac{\overline{P}_{ij}-\underline{P}_{ij}}{2} + \left| Y_{ij} - \frac{\overline{P}_{ij}+\underline{P}_{ij}}{2}\right| = \\ 
    \max_i \left\{ \sum\limits_{j: Y_{ij}\leq \widetilde{P}_{ij}} (\overline{P}_{ij} - Y_{ij}) + \sum\limits_{j: Y_{ij}> \widetilde{P}_{ij}} ( Y_{ij} - \underline{P}_{ij}) \right\} 
    \eqcolon \rho(M,Y)
    \leq \\
    \max\limits_i\!\! \left\{ \sum\limits_{j: Y_{ij}\leq \widetilde{P}_{ij}} (\overline{P}_{ij} - Y_{ij}) + \!\!\!\!\!\!\!
    \sum\limits_{j: Y_{ij}> \widetilde{P}_{ij}} ( Y_{ij} - \underline{P}_{ij})                  + \!\!\!\!\!\!\!
    \sum\limits_{j: Y_{ij}> \widetilde{P}_{ij}} (\overline{P}_{ij} - Y_{ij}) + \!\!\!\!\!\!\! \sum\limits_{j: Y_{ij}\leq \widetilde{P}_{ij}} ( Y_{ij} - \underline{P}_{ij}  )
    \right\}    =   \\
    \max_i \sum\limits_j (\overline{P}_{ij} - \underline{P}_{ij}) \eqcolon \rho(M).
\end{multline}
With the given definitions, we now have that for any $Y\in\MCs(M)$:
\begin{equation} \label{eq:sup:PMDPeps}
    \MCs(M) \subseteq \bigl\{ Y+E \given \|E\|_\infty \leq \rho(M, Y) \bigr\} \subseteq \bigl\{ Y+E \given \|E\|_\infty \leq \rho(M) \bigr\} .
\end{equation}
For convenience, we restate Theorem~\ref{thm:bound}:
\begin{theorem}\label{theorem: gap}
    Let $P_C=P_{\pi^C}$ be scrambling, $w_C$ be the corresponding Perron vector, and $r_C=r_{\pi^C}$,
    where $\pi^C$ denotes the myopic policy.
    Then, the optimal policy outperforms the myopic policy by no more than the following bound:
    \begin{equation}%
        \frac{g^* - g^C}{\overline{r}} \leq \frac{1}{1-\tau_1(P_C)} \rho(M, P_C).
    \end{equation}
    where $\rho(M,\cdot)$ is defined in \eqref{eq:Einfty}, and $\overline{r} = \max_{s,a} R(s,a)$ is the maximal available reward.
\end{theorem}
\begin{pf}
    We apply Theorem~\ref{thm:seneta} to derive
    \begin{multline*}
        g^* - g^C = {w^*} r_* - {w_C}\trans r_C = \max_{\pi\in \text{MR}} \{ w_\pi\trans r_\pi - w_C\trans r_C \}= \\
        \max_\pi \{(w_\pi\trans - w_C\trans) r_C +  w_\pi\trans (r_\pi  - r_C) \}= \\
        \max_\pi \{(w_\pi\trans - w_C\trans) r_\pi +  w_C\trans (r_\pi  - r_C)\} \leq \\
        \max_\pi \{\|w_\pi - w_C\|_1\} \|r_C\|_\infty = \max_\pi\{ \|w_\pi - w_C\|_1 \}\, \overline{r} \\
        \leq 
        \overline{r} \, \frac{1}{1-\tau_1(P_C)}  \rho(M, P_C),
    \end{multline*}
    where we used $\|r_C\|_\infty=\overline{r}$ and $(r_\pi-r_C)\leq 0$ $\forall\pi$.
\end{pf}
\begin{corollary}%
    If $P(s' \given s, a) = P(s' \given s)$ for all $s,s',a$, i.e.\ if all pages of $P(:)$ are equal, then $g^C=g^*$.
\end{corollary}
\begin{pf}
    Consider again~\eqref{eq:Einfty:sup} and the definitions just before. If all pages of $P(:)$ are equal, then we have $\MCs(M)=\{P(1)\}$ and hence $\underline{P}_{ij}=\overline{P}_{ij}=Y_{ij}$ $\forall i,j$, and thus $\rho(M,Y)\equiv0$.
\end{pf}
\subsection{Likelihood of uncontrolled and controlled MDP}\label{ssec:sup:LR}
Recall that the model of the uncontrolled and controlled MDP environment is denoted as $M_0$ and $M_1$,  
respectively.  

\paragraph*{Likelihood of $M_0$} Because we assume that each state 
is independent of action taken,
the probability of state sequences under $M_0$ is fully parametrized by 
$\theta_0=(p(1 \mid 1), p(1 \mid 2), \cdots, p(N \mid N))$, where $p(1\mid1)=P(s_{t+1}=1 \mid s_{t} =1)$ and so on. 
We assume that $P(s_0=s)=1/N$ for all $s$.
Then, the probability of observing $\Obs$ is
\begin{equation}
    P(\Obs\mid\; \theta_0) = \frac{1}{N}p(s_1 \mid s_0)p(s_2 \mid s_1)\dotsm p(s_T \mid 
    s_{T-1})=\\
    \frac{1}{N} \prod_{s'=1}^N \prod_{s=1}^N p(s' \mid s)^{m'(s', s)}.
\end{equation}
This likelihood is maximized at the maximum-likelihood estimate 
$\hat\theta_0=(\hat{p}(1 \mid 1),\dotsc)$ with
\[ 
\hat{p}(s' \mid s) = \begin{cases}
    \frac{m'(s', s)}{n'(s)} & \text{if } n'(s)\geq1\\
    \text{undefined} & \text{else}
\end{cases}
\]
and hence we get
\[
\ell_0 = P(\Obs\mid\; \hat\theta_0) =  
\frac{1}{N}\prod_{s'=1}^N \prod_{s=1}^N \left(\frac{m'(s',s)}{n'(s)}\right)^{m'(s',s)}.
\]
Note that the undefined values do not appear in this computation, so $\ell_0$ is well-defined.  

\paragraph*{Likelihood of $M_1$} In order to parametrize the probability of state sequences in an MDP, the parameter vector $\theta_1=\bigl(p(1|1,1), p(1|2,1),\dotsc,p(N| N,A)\bigr)$ needs to contain all the transition probabilities
\[
p(s' | s,a) = P( s_t = s'\mid s_{t-1}=s, a_{t-1}=a).
\]
We again assume that $P(s_0=s)=1/N$ for all $s$, and we can then write the probability of observing $\Obs$ given a parameter vector $\theta_1$ as
\begin{equation}
    P(\Obs\given\theta_1) = 
    \frac{1}{N}
    p(s_1| s_0,a_0) p(s_2| s_1,a_1) \dotsm p(s_T | s_{T-1},a_{T-1}) = %
    \frac{1}{N} \prod_{s'=1}^N\prod_{s=1}^N\prod_{a=1}^A
    p(s'| s,a)^{m(s',s,a)}.
\end{equation}
This is maximized at the maximum-likelihood estimate $\hat\theta_1$ with
\begin{equation}
    \hat{p}(s'|s,a) = \begin{cases}
        \frac{m(s',s,a)}{n(s,a)} & \text{if } n(s,a)\geq1\\
        \text{undefined} & \text{else}
    \end{cases}
\end{equation}
Hence, we can compute the likelihood of $M_1$ as 
\begin{equation}
    \ell_1 = \frac{1}{N} \prod_{(s,a): n(s,a)\geq1} \prod_{s'=1}^N 
    \left( \frac{m(s',s,a)}{n(s,a)} \right)^{m(s',s,a)}.
\end{equation}

\subsection{Proof of Theorem \ref{thm:type2}}\label{ssec:sup:type2}
For the given event $\{L \le t \,\, \text{up to time $T$}\}$, the algorithm ${\cal A}_0$ is employed throughout the interval $[0,T]$. Therefore, 
    $\beta(T)$ is upper bounded by $\P(L \le t \,|\, {\cal A}_0, \, H_1)$. Employing the fixed algorithm ${\cal A}_0$ implies that the
combined system of MDP and exploration policy generates a homogeneous Markov chain on the space $\Omega = {\cal S} \times {\cal A}$ with transition matrix $Z$:
\be\label{def:T}
Z(\omega' | \omega) = \pi^{(E)}_{a'}(s') \, P(s' | s, a), \quad \omega = (s,a), \quad \omega' = (s',a').
\ee
$Z$ will be assumed irreducible with Perron vector $w_Z = \{w_Z(s,a)\} = \{\pi^{(E)}_{a}(s) \, w_I(s)\}$.
Given a sequence of observations $\{(s_0,a_0), (s_1,a_1), \dots, (s_n,a_n)\}$, define the counts
\be\label{def:m''}
m''(s',a',s,a) = {\rm card}\{ t : (s_t,a_t) = (s',a'), \, (s_{t-1},a_{t-1}) = (s,a) \}.
\ee
The estimator for $Z$ is
\be\label{def:wideT}
\widehat{Z}(\omega' \,|\, \omega) = \frac{m''(\omega', \omega)}{\sum_{\omega'} m''(\omega', \omega)}; \quad
\omega=(s,a), \, \omega'=(s',a'),
\ee
and the previously defined counts are obtained from $m''$ using partial sums:
\bee
m(s', s ,a) &= \sum_{a'=1}^A m''(s', a', s ,a), \,\quad
n(s,a) = \sum_{s'=1}^N m(s',s,a), \\
m'(s',s) &= \sum_{a=1}^A m(s',s,a), \,\quad
n'(s) = \sum_{s'=1}^N m'(s',s).
\eee
These define estimators for the transition matrices $P$ and likelihood ratios:
\be
\widehat{P}(s' | s, a) &= \frac{m(s', s ,a)}{n(s,a)}, \quad n(s,a) > 0; \\
\widehat{P}(s' | s) &= \frac{m'(s',s)}{n'(s)}, \quad n'(s) > 0; \\
\ln \widehat{l_0} &= \sum_{s',s \, : \, m'(s',s) > 0} m'(s',s) \, \ln \widehat{P}(s'|s); \\
\ln \widehat{l_1} &= \sum_{s',s,a \, :\, m(s',s,a) > 0} m(s',s,a) \, \ln \widehat{P}(s'|s,a).
\ee
The LR test statistic is $L = 2 \ln \widehat{l_1} - 2 \ln \widehat{l_0}$, and we define
\begin{equation}
    G = \lim\limits_{n \rightarrow \infty} \frac{1}{n} L %
    = 2\sum\limits_{s',s,a  \in {\cal R}_1} w_Z(s,a) P(s' | s, a)    \ln P(s' | s, a) 
    - 2\sum\limits_{s',s \in {\cal R}_0} w_I(s) P(s' | s) \ln P(s' | s),
\end{equation}
where ${\cal R}_0 = \{(s',s) \,:\, P(s' | s) > 0\}$, ${\cal R}_1 = \{(s',s,a) \,:\, P(s' | s, a) > 0\}$.

\begin{lemma}\label{lem:G}
    \begin{equation}
    G \ge \frac{\varepsilon \, \theta^2}{4 A} \, \underline{w_I}.
    \end{equation}
\end{lemma}

Lemma \ref{lem:G} implies that $G > 0$ under the hypothesis $H_1$, and therefore $L \rightarrow \infty$ as $n \rightarrow \infty$.
We use large deviation techniques \cite{DZ} to control the rate of convergence,  and hence prove exponential decay of $\beta$.
Define  $g = (\varepsilon \, \theta^2/8 A) \, \underline{w_{I}}$ and 
\begin{align*}
\ln {l_0} &= \sum_{s',s \in {\cal R}_0} m'(s',s) \, \ln P(s'|s), &
\ln {l_1} &= \sum_{s',s,a \in {\cal R}_1} m(s',s,a) \, \ln P(s'|s,a).
\end{align*}
Noticing that $\ln \widehat{l_1} \ge \ln {l_1}$,
and taking $n \ge t/g$, we can write
\be\
\beta(n) & \le  \P\left(G - \frac{2}{n} \ln \frac{{l_1}}{ {l_0}} + 
\frac{2}{n} \ln \frac{\widehat{l_0}}{ {l_0}} \ge g\right) %
= \P\left(V_1 + V_2 + V_3 \ge g\right) \nonumber \\
& \le  \P\left(V_1 \ge \frac{g}{3}\right) + \P\left(V_2 \ge \frac{g}{3}\right) + \P\left(V_3 \ge \frac{g}{3}\right),
\ee
where
\be
V_1 &= 2 \sum_{s',s,a \in {\cal R}_1} \widehat{w_Z}(s,a) \, \left[P(s' | s,a) - \widehat{P}(s'|s,a) \right] \, 
\ln \frac{P(s' | s,a)}{P(s' | s)}, \nonumber \\
V_2 &= 2 \sum_{s',s,a \in {\cal R}_1} P(s' | s,a) \, \left[w_Z(s,a) - \widehat{w_Z}(s,a) \right] \, \ln \frac{P(s' | s,a)}{P(s' | s)}, \nonumber \\
V_3 &= 2 \sum_{s',s \,: \,m'(s',s) > 0}  \widehat{w_I}(s) \, \widehat{P}(s'|s) \, \ln \frac{\widehat{P}(s'|s)}{P(s' | s)}, \nonumber \\
\widehat{w_Z}(s,a) &= \frac{n(s,a)}{n} , \quad
\widehat{w_I}(s) = \frac{n'(s)}{n} \quad \text{(estimators for Perron vectors)}. \nonumber 
\ee

\begin{lemma}\label{lem:LDP}
    For any $y >0$,
    \bee
    \overline{\lim}_{n \rightarrow \infty} \frac{1}{n} \ln \P\left(V_j \ge y\right) &\le 
    \begin{cases}
        - y^2 \,P_{min}^2 \, \theta^{-2} \,N^{-1} \,A^{-1} /2 & j=1 \\
        - y^2 \, P_{min}^2 \,(1 - \tau_1(P))^2 \,\theta^{-2}/8 & j=2 \\
        - y/2 & j=3 \end{cases}
    \eee
\end{lemma}

\medskip
\par\noindent{\em Proof of Theorem \ref{thm:type2}:}
substitute $y=g/3$ in Lemma \ref{lem:LDP} and define
\bee
v = \frac{\varepsilon^2 \, \theta^2 \, P_{min}^2 \, \underline{w_{I}}^2}{ 2 (24 A)^2} \, \min \bigg\{\frac{1}{A N} , \frac{(1 - \tau_1(P))^2}{4} \bigg\}.
\eee
Then,
\be\label{lim1}
\overline{\lim}_{n \rightarrow \infty} \frac{1}{n} \ln \beta \le
\max_{j=1,2,3} \bigg\{ \overline{\lim}_{n \rightarrow \infty} \frac{1}{n} \ln \P\left(V_j \ge \frac{g}{3}\right) \bigg\}
\le - v.
\ee
For $\delta > 0$, (\ref{lim1}) implies that for all $n$ sufficiently large
\bee
\frac{1}{n} \ln \beta \le - v + \delta,
\eee
and therefore
\bee
e^{n (v - 2 \delta)} \beta \le e^{- n \delta} \rightarrow 0 \quad \text{as $n \rightarrow \infty$}.
\eee
Since this holds for any $\delta > 0$, it follows that $\kappa^* \ge v$.

\medskip
\par\noindent{\em Proof of Lemma \ref{lem:G}}:
$G$ can be written as
\be
G = 2 \sum_{s,a} \pi^{(E)}_{a}(s) \, w_I(s) \, D(P( \cdot | s,a) \,||\, P( \cdot | s)),
\ee
where $D$ is relative entropy, $P( \cdot | s)$ represents the distribution $\{p( s' | s)\}$ restricted to $s'$ such that
$s',s \in {\cal R}_0$, and similarly for $P( \cdot | s,a)$.
Pinsker's inequality implies that for any $s',s,a \in {\cal R}_1$
\be\label{G-eqA2}
G & \ge  \pi^{(E)}_{a}(s) \, w_I(s) \,  | p( s' | s,a) - p( s' | s) |^2.
\ee
From $\theta=\max_{i,j,a,b}  |P(j | i, a) - P(j | i, b)|$ it follows that $\max_{s',s,a} |P(s' | s, a) - P(s' | s)| \ge \theta/2$, so choosing
$s',s,a$ in (\ref{G-eqA2}) to be these maximizers, we get
\bee
G \ge \pi^{(E)}_{a}(s) \, w_I(s) \, \frac{\theta^2}{4} \ge \frac{\varepsilon \, \theta^2}{4 A} \, \underline{w_I}.
\eee

\medskip
\par\noindent{\em Proof of Lemma \ref{lem:LDP}:}
define ${\cal K} = \{(\omega,\omega') \subset \Omega \times \Omega \,:\, Z(\omega' | \omega) > 0\}$,
and let ${\cal M}({\cal K})$ denote the set of stationary probability measures on ${\cal K}$.
The large deviation rate function \cite{DZ} for the pair empirical measure on the Markov chain (\ref{def:T})
is the map $\phi_2 : {\cal M}({\cal K}) \rightarrow \mathbb{R} \cup \{  \infty \}$ defined by 
\be\label{def:phi}
\phi_2(Q) = \sum_{\omega,\omega' \in {\cal K}} Q(\omega,\omega') \, 
\ln \frac{Q(\omega,\omega')}{Q_1(\omega) \, Z(\omega' | \omega)} , \quad Q_1(\omega) = \sum_{\omega'} Q(\omega,\omega').
\ee
Therefore, for any set $\Gamma \subset {\cal M}({\cal K})$
\bee
\overline{\lim}_{n \rightarrow \infty} \frac{1}{n} \, \ln \P( \widehat{Z} \in \Gamma) \le - \inf_{Q \in \overline{\Gamma}} \phi_2(Q).
\eee
Lemma \ref{lem:LDP} follows by estimating the infimum of $\phi_2$ over the sets defined by the three events $\{V_j > y\}$.
First, note that $|P(s' | s,a) - P(s' | s)| \le \theta$ for all $s',s,a$, and thus
\bee
\bigg| \ln \frac{P(s' | s,a)}{P(s' | s)} \bigg| \le \max_{\pm} \bigg| \ln \frac{P(s' | s,a)}{P(s' | s,a) \pm \theta} \bigg| \le \frac{\theta}{P_{min}}.
\eee
For $j=1$, we have
\bee
V_1 &= \sum_{s',a',s,a \in {\cal R}_1} \widehat{w_Z}(s,a) \, \left[P(s',a' | s,a) - \widehat{P}(s',a' | s,a) \right] \, 
\ln \frac{P(s' | s,a)}{P(s' | s)} \\
& \le  \frac{\theta}{P_{min}} \, \sum_{\omega', \omega} \widehat{Z}_1(\omega) \, |Z(\omega' | \omega) - \widehat{Z}(\omega' | \omega)| %
 \le  \frac{\theta \sqrt{N A}}{P_{min}} \, 
\left(\sum_{\omega', \omega} \widehat{Z}_1(\omega) \, |Z(\omega' | \omega) - \widehat{Z}(\omega' | \omega)|^{2}\right)^{1/2} \\
& \le  \frac{\theta \sqrt{N A}}{P_{min}} \, \left(2 \phi_2(\widehat{Z})\right)^{1/2}.
\eee
Therefore,
\bee
\P\left(V_1 \ge y\right) \le \P\left(\phi_2(\widehat{Z}) \ge \frac{y^2 \, P_{min}^2}{2 \theta^2 N A}\right),
\eee
which immediately implies the result. For $j=2$, we use the large deviation rate function for the singlet empirical measure \cite{DZ}:
\bee
\phi_1(Q_1) &= \sup_{u > 0} \sum_{\omega'} Q_1(\omega') \ln \frac{u(\omega')}{\sum_{\omega} u(\omega)Z(\omega' | \omega)} \\
& \ge  \sum_{\omega'} Q_1(\omega') \ln \frac{Q_1(\omega')}{\sum_{\omega} Q_1(\omega)Z(\omega' | \omega)} \\
& \ge  \frac{1}{2} \, \| Q_1 - Q_1 Z \|_1^2 \\
& \ge  \frac{1}{2} \, (1 - \tau_1(Z))^{2} \, \| Q_1 - w_Z \|_1^2.
\eee
Therefore,
\bee
V_2 & \le  \frac{2 \theta}{P_{min}} \,  \sum_{s',s,a \in {\cal R}_1} P(s' | s,a) \, | w_Z(s,a) - \widehat{w_Z}(s,a) | \\
& = \frac{2 \theta}{P_{min}} \,  \sum_{s,a \in {\cal R}_1}  | w_Z(s,a) - \widehat{w_Z}(s,a) | \\
& \le  \frac{2 \theta}{P_{min}} \, (1 - \tau_1(Z))^{-1} \, \sqrt{2 \phi_1(\widehat{w_Z})},
\eee
and hence
\bee
\P\left(V_2 \ge y\right) \le \P\left(\phi_1(\widehat{w_Z}) \ge \frac{y^2 \, P_{min}^2 \, (1 - \tau_1(Z))^2}{8 \theta^2}\right),
\eee
and the result follows after noting that $\tau_1(Z) = \tau_1(P)$, where the ergodicity coefficient of $P$ is defined by
\be\label{def:tau}
\tau_1(P) = \sup_{\{z \neq 0 \,|\, \sum_{s,a} z(s,a) = 0 \}} \frac{\sum_{s'} |\sum_{s,a} z(s,a) P(s' | s,a)|}{\sum_{s,a} |z(s,a)|}.
\ee
Finally, for $j=3$, we use the smaller chain $P(s' | s)$ on ${\cal S}$ and note that $V_3 = 2 \phi_N(\widehat{P})$, where
$\phi_N$ is the large deviation rate function for $P(s' | s)$ corresponding to (\ref{def:phi}). The result follows immediately.

\subsection{Proof of Corollary~\ref{cor:regret}}\label{ssec:sup:regret}
Assume the environment is a controlled MDP, i.e., the null hypothesis is wrong.
    The statistic $L$ may switch back and forth between the regions $\{L \le t\}$ and $\{L > t\}$ throughout the time interval $[0,T]$.
    Let ${\cal T}_0$ denote the collection of intervals during which the orchestrator uses the algorithm ${\cal A}_0$,
    and let ${\cal T}_1$ denote the collection of intervals during which the orchestrator uses the algorithm ${\cal A}_1$.
    By Theorem~\ref{thm:type2}, there are constants $0<\kappa<\kappa^*$ and $C_\kappa$ such that
    the probability of a sojourn of length $s$ in the region $\{L \le t\}$ is bounded by $C_\kappa e^{-\kappa s}$. 
    It follows that the expected duration of such a sojourn is bounded by 
    $\sum_{s=1}^{\infty} s C_\kappa e^{-\kappa s} = C_\kappa'$.
    Then,
    \bee
    \E\{R(T)\} &= \E\left\{\sum_{t \in {\cal T}_0} r(s_t,a_t^*) - r(s_t,a_t)\right\} 
    +  \E\left\{\sum_{t \in {\cal T}_1} r(s_t,a_t^*) - r(s_t,a_t)\right\}\\
    &\le  \E\left\{ \sum_{t \in {\cal T}_0} \bar{r} \right\} 
    +  \E\left\{\sum_{t \in [0,T]} r(s_t,a_t^*) - r(s_t,a_t)\right\}\\
    & \le  \bar{r} \, C_\kappa' \,  \E\left\{|{\cal T}_0|\right\} + R^1_\text{c}(T).
    \eee
    From the proof of Theorem~\ref{thm:type2} it follows that $L \rightarrow \infty$ as $T \rightarrow \infty$ 
    for a controlled MDP-environment when the algorithm ${\cal A}_1$ is used. When the test statistic enters the
    region $\{L > t\}$ and the orchestrator switches to ${\cal A}_1$,  
    there is a non-zero probability (again for a controlled MDP-environment) that
    $L > t$ at every subsequent step, implying that the orchestrator does not switch back to ${\cal A}_0$. Letting $q$ denote this
    probability, it follows that the expected number of times that the orchestrator switches back to ${\cal A}_0$
    before eventually remaining with algorithm ${\cal A}_1$ is $q^{-1}$. Therefore,
    $\E\left\{|{\cal T}_1|\right\} = q^{-1}$, and since $|{\cal T}_0| \le |{\cal T}_1| + 1$, this leads to the bound
    \bee
    \E\{R(T)\} & \le  \bar{r} \, C_\kappa' \,  \E\left\{|{\cal T}_0|\right\} + R^1_\text{c}(T) \\
    & \le  \bar{r} \, C_\kappa' \,  \left(1 + q^{-1} \right) + R^1_\text{c}(T) \\
    &=  \mathrm{O}\left(R^1_\text{c}(T)\right).
    \eee
    Recall that $f(x) = \mathrm{O}(g(x))$ if there exist $M,z>0$ such that $|f(x)|\leq M|g(x)|$ for all $x\geq z$.

\end{document}